\documentclass[conference]{IEEEtran}
\usepackage{times}

\usepackage[numbers]{natbib}
\usepackage{multicol}
\usepackage[bookmarks=true]{hyperref}

\usepackage{amsmath}
\usepackage{amsfonts}
\usepackage{adjustbox}
\usepackage{comment}
\usepackage{caption}
\usepackage{subcaption}
\usepackage{xcolor}
\usepackage{wrapfig}
\usepackage{cleveref}
\makeatletter
\newcommand\notsotiny{\@setfontsize\notsotiny\@vipt\@viipt}
\makeatother

\let\oldtwocolumn\twocolumn
\renewcommand\twocolumn[1][]{%
    \oldtwocolumn[{#1}{
        \vspace{-0.5cm}
        \begin{center}
\includegraphics[height=0.6025\columnwidth]{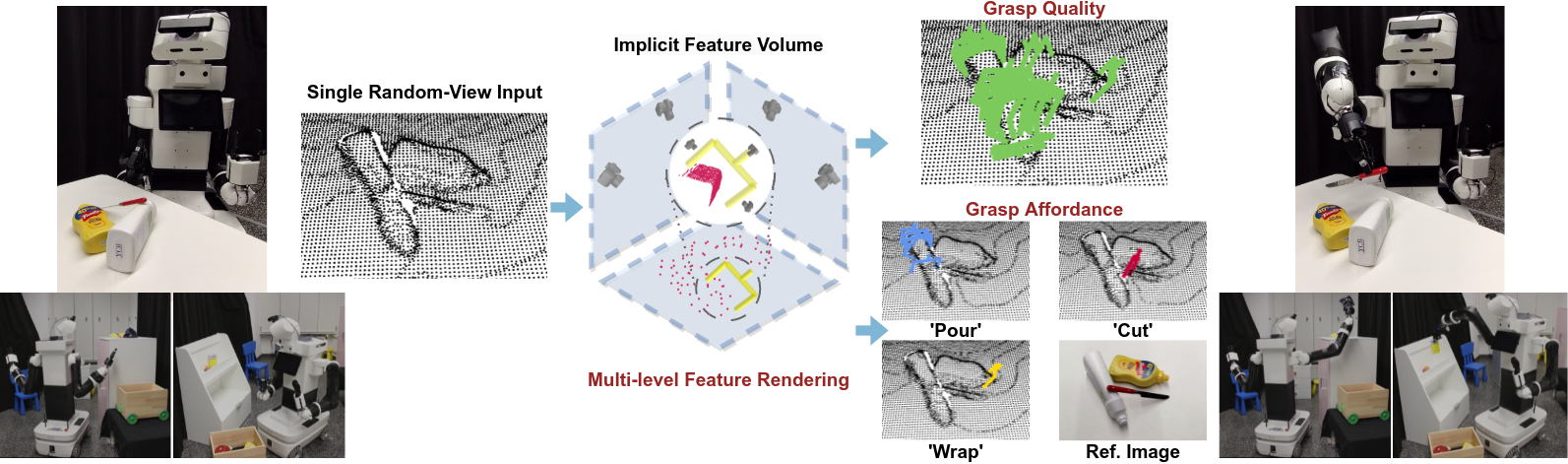}
 \vspace{-0.3cm}
           \captionof{figure}{\textbf{Grasping as rendering}. Our network, NeuGraspNet, uses a single random-view depth input, encodes the scene in an implicit feature volume, and uses multi-level rendering to select relevant features and predict grasping functions. As shown, NeuGraspNet generalizes to random-view mobile manipulation grasping scenarios. More at: \href{https://sites.google.com/view/neugraspnet}{sites.google.com/view/neugraspnet}}
           \label{fig:cover_tiago}
        \end{center}
    }]
}

\begin{document}

\title{Learning Any-View 6DoF Robotic Grasping in Cluttered Scenes via Neural Surface Rendering}


\author{%
Snehal Jauhri$^{1}$, \quad Ishikaa Lunawat$^{2}$, \quad and \quad Georgia Chalvatzaki$^{1,3,4}$\\
$^{1}$ Computer Science Dept., TU Darmstadt, Germany \\
$^{2}$ NIT Trichy, India $^{3}$ Hessian.AI, Darmstadt, Germany \\
$^{4}$ Center for Mind, Brain and Behavior, Uni. Marburg and JLU Giessen, Germany \\
}


%

\maketitle
\begin{abstract}
A significant challenge for real-world robotic manipulation is the effective 6DoF grasping of objects in cluttered scenes from any single viewpoint without needing additional scene exploration. This work re-interprets grasping as \textit{rendering} and introduces NeuGraspNet, a novel method for 6DoF grasp detection that leverages advances in neural volumetric representations and surface rendering. We encode the interaction between a robot's end-effector and an object's surface by jointly learning to render the local object surface and learning grasping functions in a shared feature space. Our approach uses global (scene-level) features for grasp generation and local (grasp-level) neural surface features for grasp evaluation. This enables effective, fully implicit 6DoF grasp quality prediction, even in partially observed scenes. NeuGraspNet operates on random viewpoints, common in mobile manipulation scenarios, and outperforms existing implicit and semi-implicit grasping methods. We demonstrate the real-world applicability of the method with a mobile manipulator robot, grasping in open cluttered spaces.
\end{abstract}
\IEEEpeerreviewmaketitle

\section{Introduction}
\label{sec:intro}

Robotic manipulation is crucial for enabling various applications such as home-assistance, industrial automation etc. A key component for manipulation is the ability to grasp objects in unstructured, cluttered spaces under partial observability. This ability would enhance the efficiency, versatility, and autonomy of robots operating in everyday environments. 
Deep learning has been crucial in making advances in robotic grasping~\cite{mahler2017dex,mahler2018dex,morrison2020learning,chalvatzaki2020orientation} by training networks using simulation data and transferring to the real world. However, 6DoF grasping in the wild, i.e., grasping in the SE(3) space from \textit{any viewpoint} remains a challenge~\cite{kroemer2021review,newbury2022deep,Platt_review}. Embodied AI agents, e.g., mobile manipulation robots~\cite{jauhri2022robot,hu2023causal}, are expected to perform manipulation tasks similar to humans; humans can leverage geometric information from limited views and mental object models to grasp objects without exploring the whole scene. Such grasping in open cluttered spaces requires that robots, given some spatial information, e.g., 3D pointcloud data, can reconstruct the scene, understand graspable areas of objects, and detect grasps likely to succeed. Moreover, robots should reason about the grasp's affordance~\cite{gibson1977theory}, i.e., the subsequent task a grasp allows, while additionally avoiding collision with the surrounding environment.

6DoF grasping methods can be classified into methods that \textit{explicitly generate} grasp poses~\cite{Mousavian2019,wang2021graspness,Sundermeyer2021} or \textit{implicitly} classify the grasp quality of \textit{any} grasp candidate in SE(3) using a discriminative model~\cite{Pas2017,Liang2019,Huang2023}. 
The ability to assess the quality of any grasp pose implicitly is essential to applications in which grasp candidates are pre-defined due to human demonstrations~\cite{ye2023learning} or other affordance-based information~\cite{Christen_2022_CVPR, li2023contact2grasp}. Moreover, explicit generative models are difficult to combine with additional constraints since the constraints can only be applied as a post-filtering step. In implicit methods, however, the distribution of grasp candidates can be chosen and constrained \textit{before} querying the model for grasp quality. This ability is particularly useful in mobile manipulation tasks where constraints such as reachability of grasp poses are necessary.

Many existing grasping methods that use partial pointclouds either rely only on seen parts of a scene~\cite{ni2020pointnet++,Sundermeyer2021} or accumulate more information from multiple views~\cite{Breyer2020,jeng2021gdn,fang2022anygrasp}. An approach to mitigate partial observability is to use neural scene representations~\cite{Mescheder2019,Peng2020,Park2019} to learn scene completion in a continuous functional space. These representations are implicit in geometry, enabling querying arbitrary points in the scene. They also allow the learning of other geometric feature fields, making them an attractive solution for grasping~\cite{Jiang2021,dai2022graspnerf,lundell2021ddgc,sharma2023language}. However, most neural scene representations still require multi-view information or overfit to specific objects/scenes. 

This work investigates how to effectively leverage geometric and surface information about objects in \textit{any scene} perceived from \textit{any partial view} to detect high-fidelity 6DoF grasps. We propose a novel method, \textbf{NeuGraspNet}\footnote{The acronym hints at the novel view of grasping as neural surface rendering, and a wordplay for the German word `neu' that means new.} for 6DoF grasp detection building on advances in neural surface rendering~\cite{niemeyer2020differentiable,Oechsle2021}. 
We use a learned implicit scene representation to reconstruct and render the scene globally and effectively sample grasp candidates, even in occluded regions. Moreover, we argue that local geometric object features are essential for understanding the complementarity between the robot's end-effector and the object's surface for predicting grasp success. Thus, we treat \textit{grasping as local neural surface rendering}. We learn shared local features that encode the response of an object part to a grasping pose, enabling fully implicit grasp quality evaluation in SE(3). To evince the benefit of NeuGraspNet, we show superior performance compared to representative implicit and semi-implicit baselines. We also demonstrate real-world applicability via sim-to-real transfer to a mobile manipulator robot grasping in open spaces.

In summary, our contributions are: (i) a new neural grasp quality network (NeuGraspNet) that operates on single viewpoints, \textit{generalizes to any scene}, and is \textit{fully implicit} in both SE(3) and scene/object geometry, (ii) a new structured method for sampling grasp candidates in occluded parts of the scene using global (scene-level) rendering, (iii) the re-interpretation of \textit{grasping as surface rendering} for extracting local geometric features that capture the interaction between the robot's end-effector and the local object-geometry per grasp candidate. 


\section{Background \& Related Work}
\label{sec:rel_work}

\begin{figure*}[t!]
  \centering
  \includegraphics[width=0.79\linewidth]{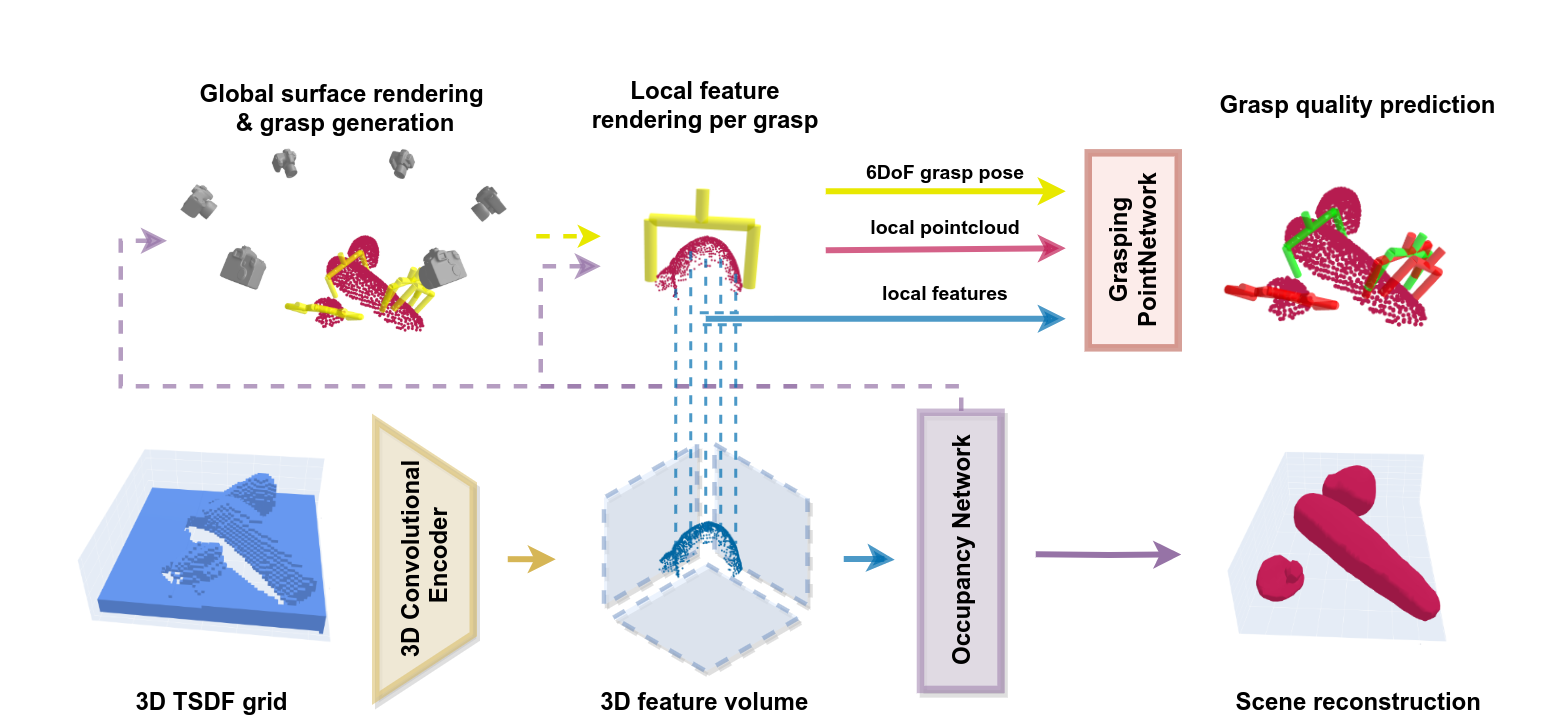}
  \caption{\textbf{NeuGraspNet}: A single-view 3D Truncated Signed Distance Field (TSDF) grid is processed through a convolutional occupancy network to reconstruct the scene (cf.~\Cref{subsec:scene}). The occupancy network is used to perform global, scene-level rendering. The rendered scene is used for grasp candidate generation in SE(3) (cf.~\Cref{subsec:gpg}). We re-interpret \textit{grasping as rendering} of local surface points and query their features from the \textit{shared} 3D feature volume. Local points, their features, and the 6DoF grasp pose are passed to a Grasping PointNetwork to predict per grasp quality (cf.~\Cref{subsec:local}). NeuGraspNet effectively learns the interaction between the objects' geometry and the gripper to detect high-fidelity grasps.}
  \label{fig:pipeline}
  \vspace{-0.3cm}
\end{figure*}

\subsection{Neural implicit scene representations}
Implicit scene representations are useful since they encode the scene in a continuous feature space, allowing them to be queried at any resolution. Occupancy networks~\cite{Mescheder2019,Peng2020} are neural representations for 3D reconstruction. Given a pointcloud or 3D voxel grid of the scene, they encode the scene in a feature space $\psi$ using convolutional encoders which combine global and local scene information. 
Neural surface rendering approaches~\cite{niemeyer2020differentiable,Oechsle2021,Wang2021} employ occupancy or surface models~\cite{Park2019,Wang2021} to render realistic 3D object surfaces. While some learned geometric representations require explicit 3D supervision~\cite{Mescheder2019,Park2019,Peng2020}, advances in differentiable rendering~\cite{Mildenhall2020} and surface model learning have also enabled reconstruction from multi-view 2D images only~\cite{niemeyer2020differentiable,Wang2021}. While multi-view image-based methods such as NeRFs~\cite{Mildenhall2020} are restricted to overfitting specific scenes, some initial success has been found towards generalization to arbitrary scenes \cite{yu2020pixelnerf}. 


\subsection{Grasp detection}
A comprehensive review on deep grasp synthesis is introduced in~\cite{newbury2022deep}. We refer here to a subset of works covering explicit grasp generative models, implicit (discriminative) models, and grasp affordance detection.

\noindent\textbf{Explicit grasp generation.}  Partial observability can greatly hinder explicit grasp pose generation~\cite{lundell2023constrained,alliegro2022end,wu2020grasp, mohammadi20233dsgrasp}. GraspNet~\cite{Mousavian2019} is a variational autoencoder generative model that learns a grasp distribution in SE(3), but additionally relies on an iterative grasp refinement process. ContactGraspNet\cite{Sundermeyer2021} extends this approach by relying on a gripper contact point assumption in the input pointcloud. A similar keypoint-based representation that encodes contact points and the grasp center is proposed in~\cite{chen2023kgnv2,chen2022keypoint}. 
In~\cite{gou2021rgb,wang2021graspness}, the authors show complementarity between RGB and pointcloud data while predicting graspness over piled scenes. AnyGrasp~\cite{fang2022anygrasp} shows robust performance and also proposes a method for finding temporal associations among grasps in consecutive observations. 

\noindent\textbf{(Semi-)Implicit grasp prediction.}  Implicit grasping models can predict a quality value of \textit{any} provided grasp candidate~\cite{Pas2017,cai2022real}. PointNetGPD~\cite{Liang2019} learns pointcloud features for predicting the quality of grasp candidates sampled using the method from \cite{Gualtieri2016}. 
VGN~\cite{Breyer2020} uses a scene TSDF to perform semi-implicit grasp detection (implicit in 3D position and explicit in orientation). Other semi-implicit methods have used coarse-to-fine predictions~\cite{jeng2021gdn} or Hough voting~\cite{hoang2022context}. 
EdgeGraspNet~\cite{Huang2023} follows ideas like the contact-based representation from~\cite{caivolumetric} and proposes a point-and-edge-based grasp representation producing SOTA results. Implicit geometric representations~\cite{Mescheder2019,Peng2020,Mildenhall2020} provide a valuable feature space for robot perception~\cite{khargonkar2023neuralgrasps, dai2022graspnerf}. GIGA~\cite{Jiang2021} extends VGN by sharing implicit features between scene reconstruction and grasp prediction, improving the primary grasping task. In ~\cite{weng2022neural,urain2022se}, 6DoF grasping functions are learned as cost functions for use in trajectory optimization, while CenterGrasp~\cite{chisari2023centergrasp} learns a semi-implicit grasp-object manifold. 

\noindent\textbf{Grasp affordances.} It is important to consider the task or ability \textit{afforded} by parts of objects~\cite{deng20213d, qin2019keto} when interacting with them. Several methods have extended object affordances to the affordance provided \textit{by the grasp} to enable task-oriented manipulation~\cite{kokic2020learning}. TaskGrasp~\cite{murali2020taskgrasp} follows~\cite{ardon2019learning} in using semantic knowledge graphs to generalize affordance prediction among similar objects. In~\cite{kokic2020learning, bahl2023affordances}, human activity videos are used to predict visual affordances. \citet{Chen2022implicit} jointly learn visual affordances and grasping to guide grasp candidates toward requested affordances. Recently proposed open-vocabulary methods~\cite{sharma2023language, shen2023F3RM} predict semantic affordances using language queries. These methods, however, use multi-view inputs and are restricted to overfitting to single scenes.

\noindent\textbf{Our approach.} Our goal is to perform 6DoF grasping using \textit{any-view} partial depth information. We also aim to \textit{generalize} to any scene, unlike many neural geometric methods that fit to specific scenes/objects~\cite{sharma2023language,shen2023F3RM}. Our novel method uses surface rendering at \textit{multiple levels} to learn relevant grasping features. Our method is \textit{fully implicit} in \textit{both} scene geometry and the 6 grasp dimensions since we learn the complementarity between the grasp and the scene via feature rendering. The method is demonstrated via sim-to-real transfer to real, random-view mobile manipulation scenarios. We also extend the approach to grasp affordances and show promising initial grasp affordance prediction results toward task-oriented grasping. \color{black}

\section{Learning 6DoF Grasping via Neural Surface Rendering}
\label{sec:method}

In a cluttered scene, we are given a 3D depth pointcloud $\mathbf{x}$ captured from an arbitrary viewing angle. We consider a robot equipped with a two-fingered gripper and represent grasps using the 6D gripper pose $\mathbf{g} \in$ SE(3). Given the scene information $\mathbf{x}$, the likelihood of success of a grasp is interpreted as the grasp's quality $q \in [0,1]$. 

In this setting, our model, NeuGraspNet, learns an implicit function $f^{sg}_{\theta}: \mathbb{R}^3 \to \mathbb{R}$ that represents the \textit{scene geometry} and a subsequent implicit function $h^{q}_{\omega}: \mathbb{R}^6 \to \mathbb{R}$ that evaluates the \textit{quality of candidate grasp poses} in the scene ($\theta$ and $\omega$ being jointly trainable network parameters), leading to a \textit{fully implicit} representation. Our full pipeline is depicted in Fig.~\ref{fig:pipeline}. In the following, we describe our reconstruction and scene-level rendering approach. We then explain the candidate grasp generation method that samples grasps on the rendered scene. This is followed by our implicit grasp quality evaluation function that uses local grasp-level surface rendering to learn appropriate features relevant to each grasp.

\subsection{Neural scene reconstruction}~\label{subsec:scene}
Our input pointcloud $\mathbf{x}$ only conveys partial information about the scene. To enable grasp generation in a scene-level feature space, we reconstruct the scene through an implicit scene geometry backbone, a convolutional occupancy network (ConvONet)~\cite{Peng2020}. The network first encodes the input pointcloud $\mathbf{x}$ into a feature space $\psi$. We can then implicitly query the occupancy probability of any 3D point $\mathbf{p} \in \mathbb{R}^3$ by passing the corresponding feature vector $\psi(\mathbf{x},\mathbf{p})$ through a fully-connected scene geometry decoder network $f^{sg}_{\theta}$. The network is trained using ground truth simulated occupancy values $o(\mathbf{p}) \in \{0, 1\}$ of points uniformly sampled in the scenes. Formally, the reconstruction loss is a binary cross-entropy loss between the predicted occupancy probability $\hat{o}(\mathbf{p}) = f^{sg}_{\theta}(\psi(\mathbf{x},\mathbf{p}))$ and the true occupancy $o(\mathbf{p})$:
\begin{equation}\label{eq:loss_occ}
	\mathcal{L}_{occ} = \mathcal{L}_{CE}\left(\hat{o}({\mathbf{p}}), o({\mathbf{p}})\right)
\end{equation}

Learning scene geometry as an auxiliary task for grasp quality prediction has been shown to be beneficial to grasp prediction~\cite{lundell2021ddgc,Jiang2021}. In this work, we adopt this intuitive idea and further highlight the importance of scene reconstruction not only as an auxiliary training task but as a core component of NeuGraspNet. Crucially, we use the learned implicit geometry to render the scene surface at a global level for grasp candidate generation (cf.~\Cref{subsec:gpg}) and at a local level for feature extraction per 6D grasp candidate pose (cf.~\Cref{subsec:local}).

\subsection{Scene-level rendering \& grasp candidate generation}
\label{subsec:gpg}

\begin{figure}[t!]
     \centering
     \begin{subfigure}[b]{0.15\textheight}
         \centering
         \includegraphics[width=\textwidth,trim={10cm 5cm 5cm 9cm},clip]{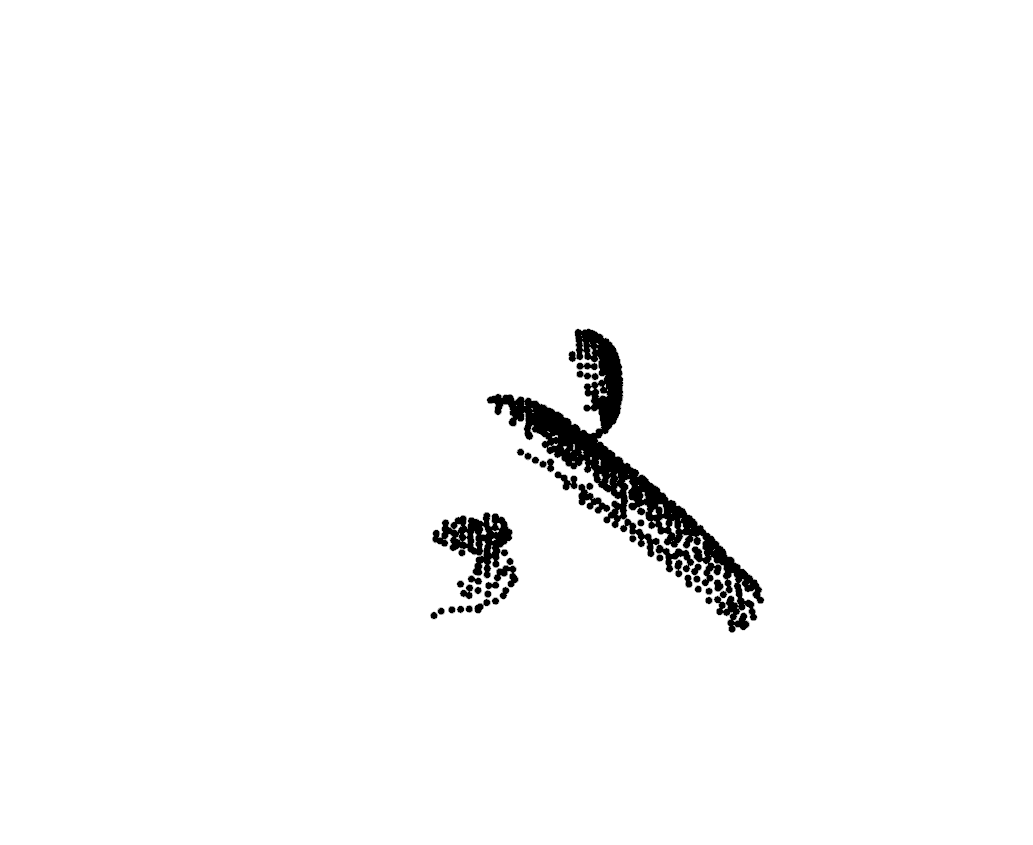}
                  \vspace{-0.4cm}
         \caption{}
     \end{subfigure}
     \begin{subfigure}[b]{0.15\textheight}
         \centering
         \includegraphics[width=\textwidth,trim={1.3cm 2cm 2cm 1.9cm},clip]{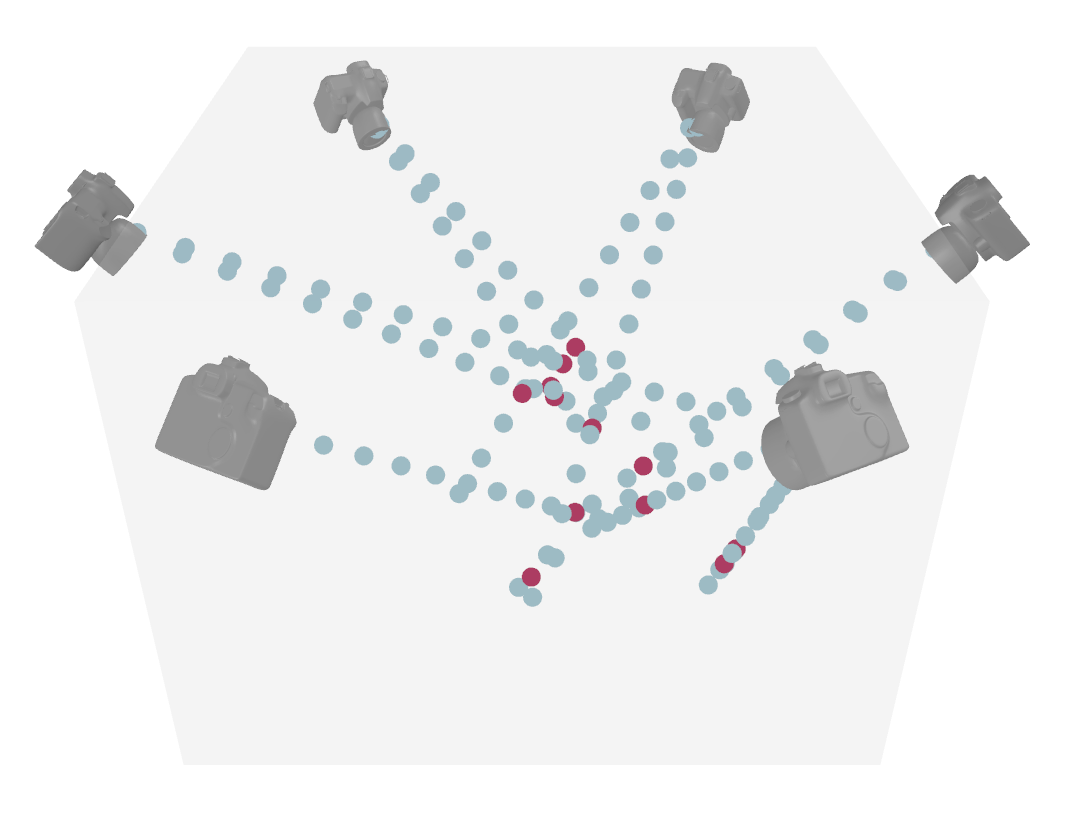}
                  \vspace{-0.4cm}
         \caption{}
     \end{subfigure}
     \begin{subfigure}[b]{0.15\textheight}
         \centering
         \includegraphics[width=\textwidth,trim={10cm 7cm 5cm 8cm},clip]{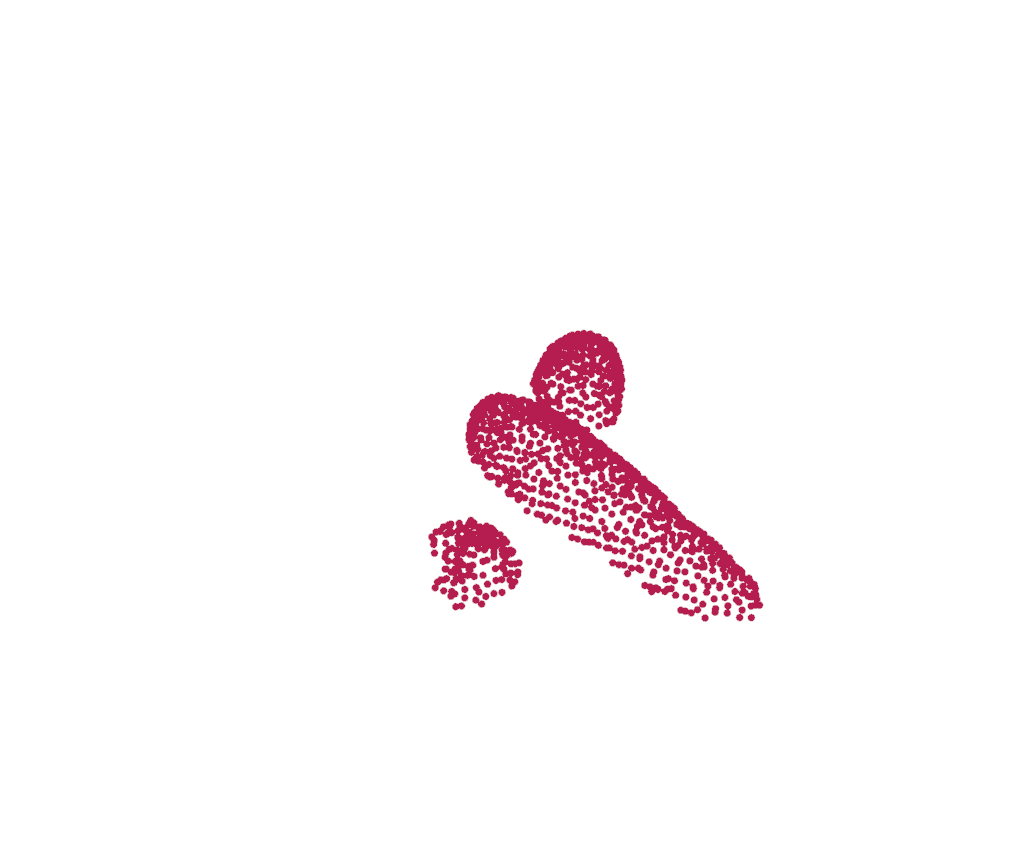}
                  \vspace{-0.4cm}
         \caption{}
     \end{subfigure}
     \begin{subfigure}[b]{0.15\textheight}
         \centering
         \includegraphics[width=\textwidth,trim={10cm 7cm 5cm 8cm},clip]{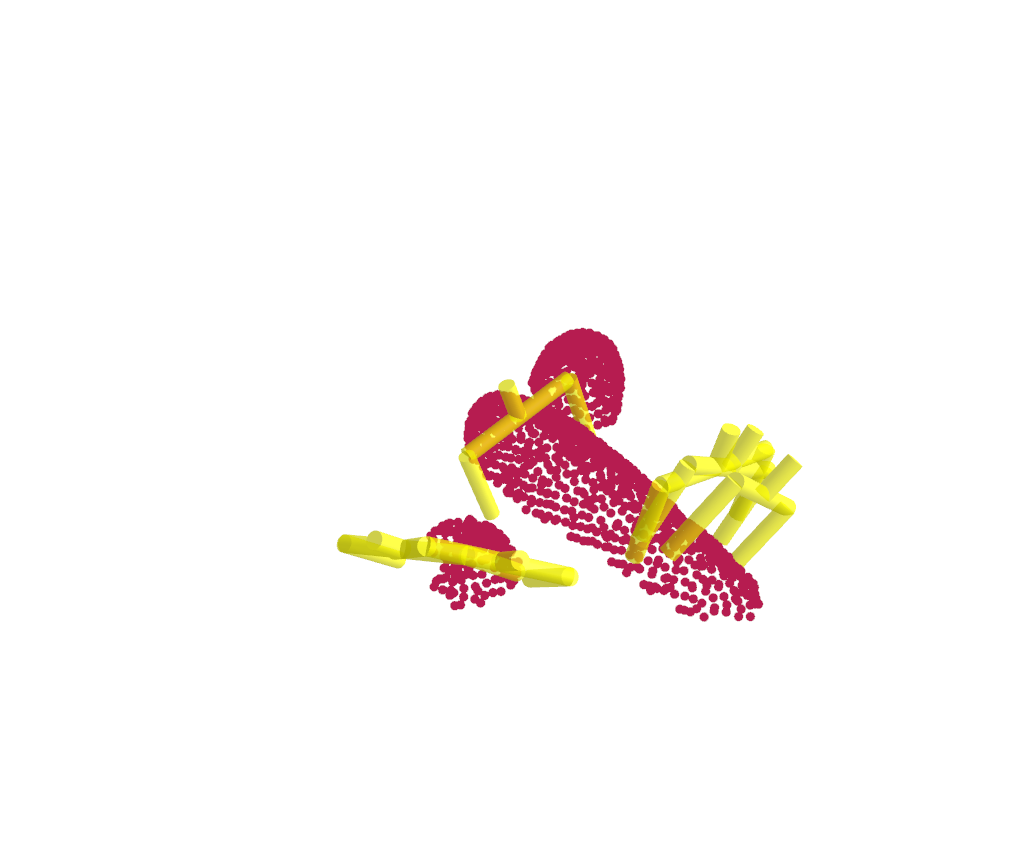}
         \vspace{-0.4cm}
         \caption{}
     \end{subfigure}
        \caption{Scene-level surface rendering: (a)~an input single-view pointcloud; (b)~surface rendering on the neural implicit geometry (grey volume) using 6 `virtual' cameras; (c)~the reconstructed surface pointcloud; (d)~sampled grasp candidates.}
        \label{fig:grasp_detec_steps}
        \vspace{-0.5cm}
\end{figure}

Using the learned occupancy representation $f^{sg}_{\theta}$, we can render the surface of the whole scene at inference time by ray-marching  $C$ ``virtual'' cameras placed in a circular path around the scene (\Cref{fig:grasp_detec_steps}). We perform surface rendering using a root-finding approach similar to \cite{niemeyer2020differentiable} and \cite{Oechsle2021}. 
Formally, for every ray $r$ emanating from each of the $C$ virtual cameras, we evaluate the occupancy network $f^{sg}_{\theta}(.)$ at $n$ equally spaced samples on the ray $\left\{\mathbf{p}_j^{r}\right\}_{j=1}^n$. The first point along a ray for which the occupancy changes from free space ($o(\mathbf{p})<0.5$) to occupied space ($o(\mathbf{p})>0.5$) is a surface point $\mathbf{p}_s$. To further refine the surface point estimation, we use an iterative secant search method along the ray (detailed in ~\cite{niemeyer2020differentiable}). After obtaining the surface point-set from all virtual cameras, we merge and downsample the point-set to arrive at a reconstructed scene pointcloud $\left\{\mathbf{p}_i\right\}_{i=1}^{M}$ as shown in Fig.~\ref{fig:grasp_detec_steps}c. We note here that other reconstruction methods, such as running marching cubes over the occupancy volume, could also be used to obtain the surface points. However, since we are only interested in the surfaces of the objects that the robot can interact with, our ray-casting procedure, which focuses on the ray's first contact with the upper surfaces of objects, reconstructs surfaces at a high resolution with fewer queries to the implicit feature volume.
\\
Our grasp detection is implicit, i.e., we can query the quality of any 6DoF grasp pose $\mathbf{g} \in$ SE(3). To generate suitable grasp proposals to discriminate upon during training, we use the sampling approach of Grasp Pose Generator (GPG)~\cite{Gualtieri2016} due to its simple yet effective nature. GPG generates grasp candidates on the input pointcloud using the point surface normals and estimating the axis of curvature of a surface. For more details, we refer to~\cite{Gualtieri2016}. 
We apply the GPG sampler on our \textit{reconstructed} surface pointcloud. This provides two benefits over the partial pointcloud. First, we can sample more grasps, since we can use the \textit{completed} scene information, i.e., sample grasps in occluded areas. Secondly, our candidates are less likely to be in collision with objects. This is because, in the partial pointcloud case, generated candidates can often intersect with objects because only parts of the object pointclouds are visible.  
A visualization of the grasps generated on our completed pointcloud can be seen in Fig.~\ref{fig:grasp_detec_steps}d.

\subsection{Local surface rendering \& learning grasping functions}\label{subsec:local}

\begin{figure}[t!]
     \centering
     \begin{subfigure}[b]{0.14\textheight}
         \centering
         \includegraphics[width=\textwidth,trim={20cm 10cm 20cm 5cm},clip]{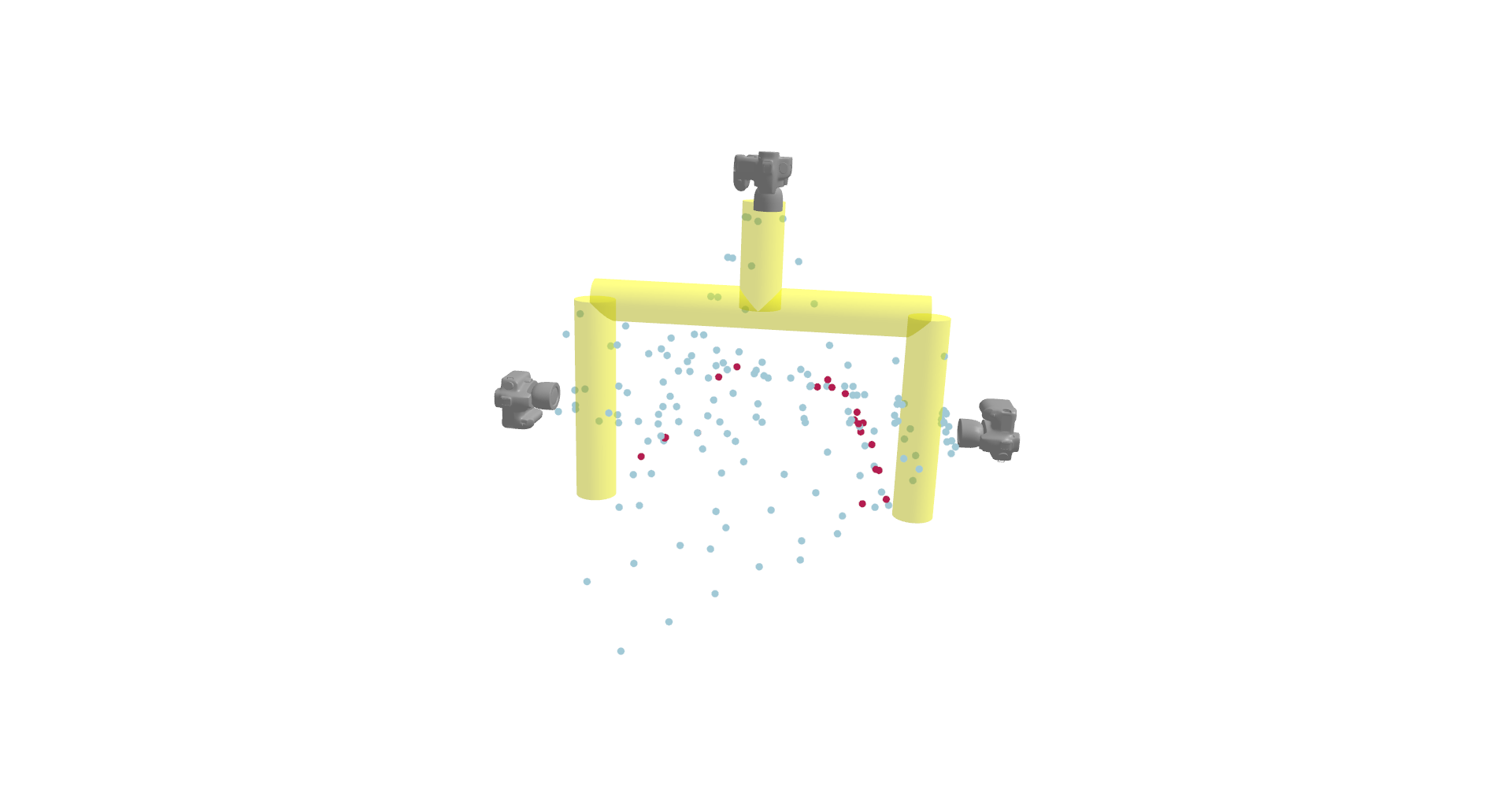}
        \vspace{-0.4cm}
         \caption{}
     \end{subfigure}
     \begin{subfigure}[b]{0.14\textheight}
         \centering
         \includegraphics[width=\textwidth,trim={20cm 10cm 20cm 5cm},clip]{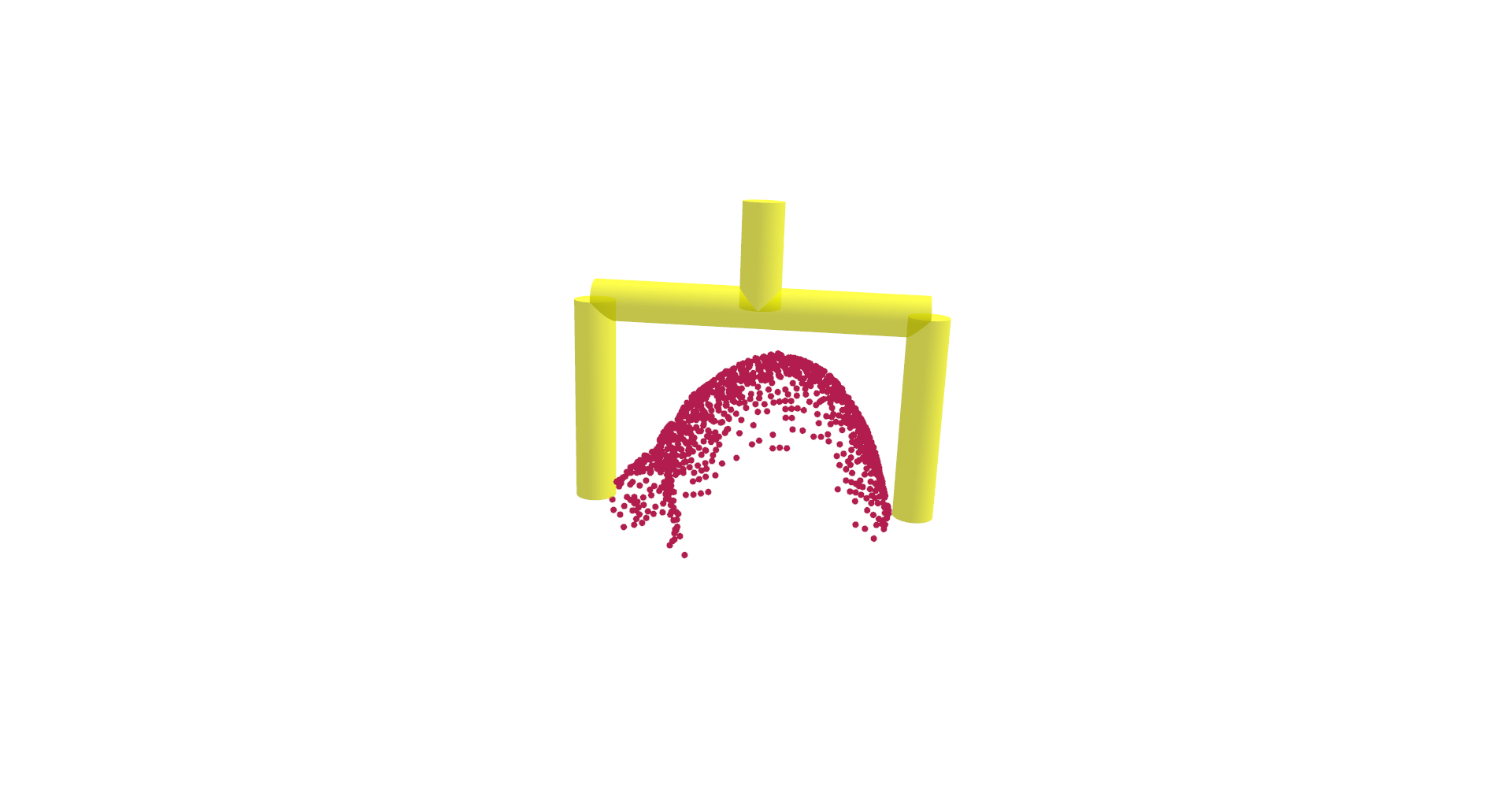}
        \vspace{-0.4cm}
                 \caption{}
     \end{subfigure}
     \begin{subfigure}[b]{0.14\textheight}
         \centering
         \includegraphics[width=\textwidth,trim={20cm 10cm 20cm 5cm},clip]{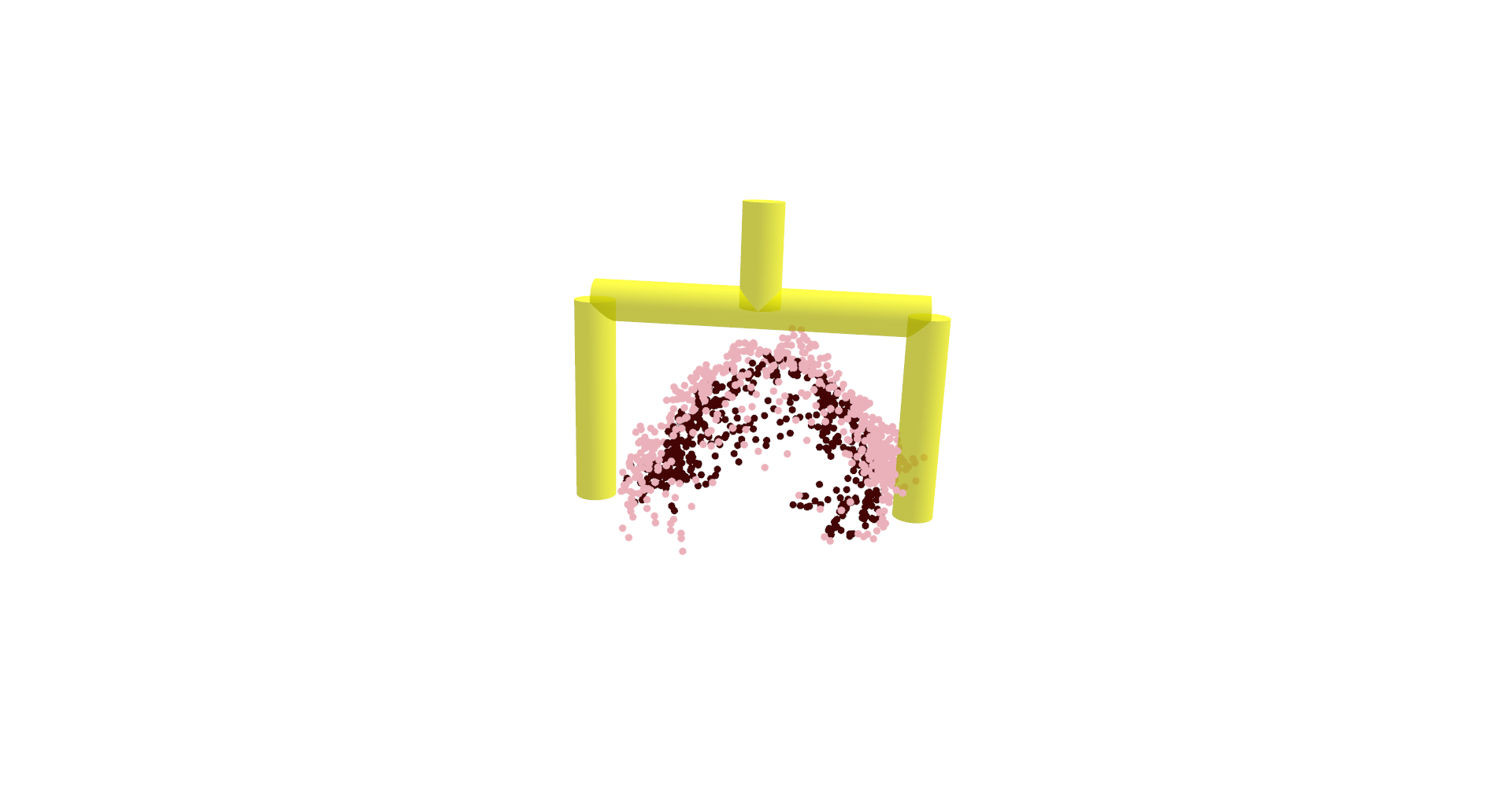}
        \vspace{-0.4cm}
                 \caption{}
     \end{subfigure}
     \begin{subfigure}[b]{0.14\textheight}
         \centering
         \includegraphics[width=\textwidth,trim={15cm 5cm 15cm 8cm},clip]{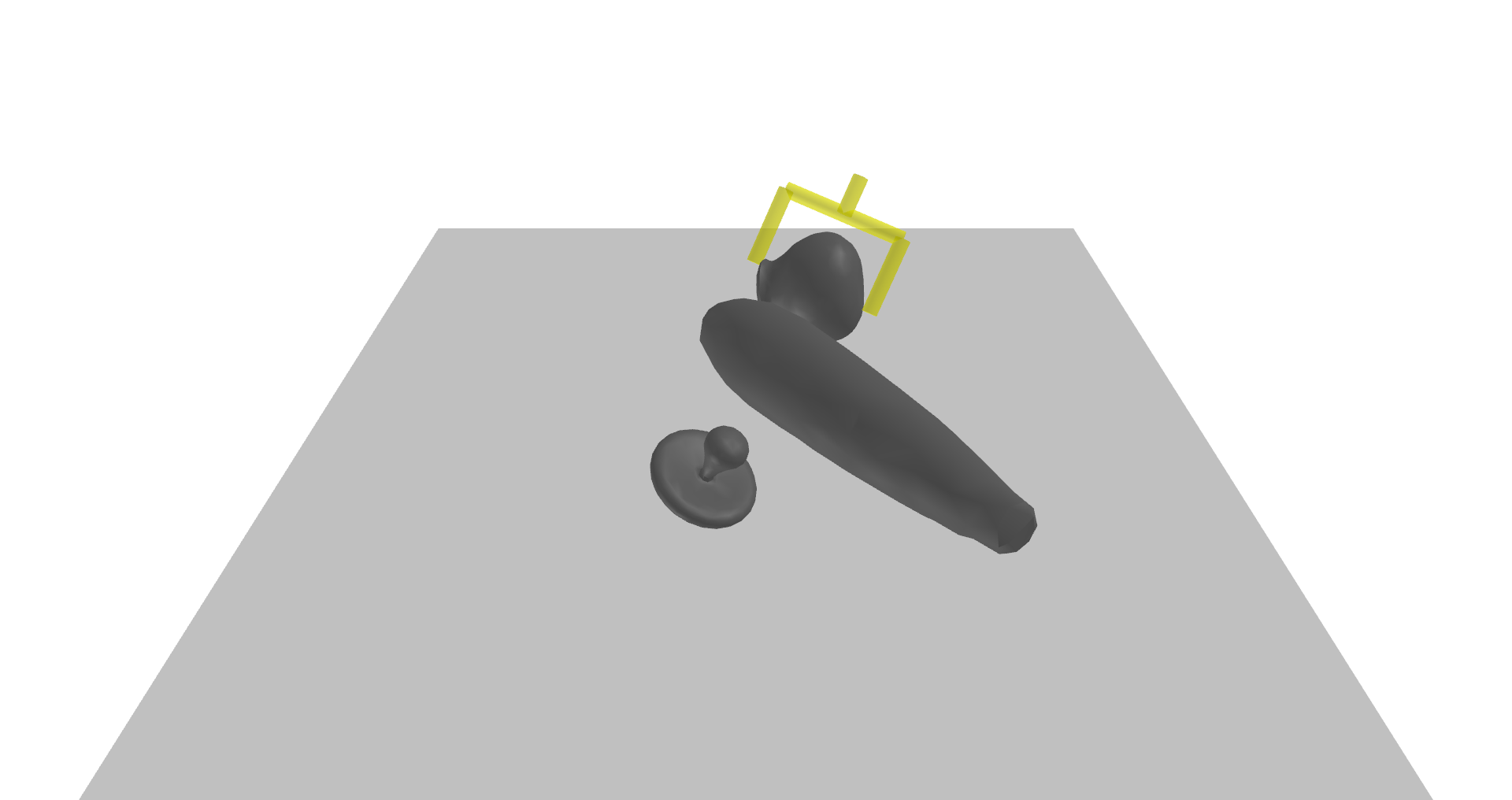}
        \vspace{-0.4cm}
                 \caption{}
     \end{subfigure}
        \caption{Local surface rendering: (a)~rendering the neural implicit geometry by ray-marching 3 `virtual' cameras at the three parts of the gripper (gripper used here only for visualization); (b)~the neural rendered surface; (c)~noisy ground-truth rendered surface used during training for local occupancy supervision (light pink points are unoccupied and dark red points are occupied); (d)~ground-truth simulated scene.}
        \label{fig:surf}
    \vspace{-0.5cm}
\end{figure}

Key to our approach is to obtain features that effectively capture the geometric interaction between any 6DoF grasp and the scene. To do so, we propose selecting the features from the scene's local surface points which are \textit{rendered} by each grasp. We hypothesize that the local surface points and their \textit{corresponding latent features} can effectively encode the geometric complementarity of object-surface and gripper to assess the quality of grasps. For each 6DoF grasp pose $\mathbf{g}$ in a scene, we use our implicit scene-geometry network $f^{sg}_{\theta}$ to render a local 3D surface point set $\left\{\mathbf{p}^{\mathbf{g}}_{i}\right\}_{i=1}^{N}$ corresponding to the grasp $\mathbf{g}$. Specifically, we place a virtual camera near each link of the gripper, i.e., near the two fingers and the base, such that the cameras point towards the inner hull of the gripper (\Cref{fig:surf}a). We thus obtain a dense local point-set and filter out spurious points too far away from the gripper links.
\\
\\
\noindent\textbf{Grasp quality prediction.} To predict the quality of a grasp $\mathbf{g}$, we use both the local surface point set of the grasp $\left\{\mathbf{p}^{\mathbf{g}}_{i}\right\}_{i=1}^{N}$ and the corresponding feature vectors in the \textit{same feature space} $\psi(\mathbf{x})$ as the scene geometry network $f^{sg}_{\theta}$. Thus, we jointly learn features $\psi$ appropriate for both scene reconstruction and grasp quality prediction~\cite{Jiang2021}. The $N$ local surface points and their features are passed through an implicit grasp quality decoder network $h_{\omega}^{q}$ to predict the grasp success probability $\hat{q}(\mathbf{g})= h_{\omega}^{q}\bigl( \left\{\mathbf{p}^{\mathbf{g}}_{i}, \psi(\mathbf{x},\mathbf{p}^{\mathbf{g}}_{i})\right\}_{i=1}^{N} \bigr)$. The grasp quality decoder uses a permutation invariant point network architecture~\cite{Qi2017}, trained using simulated ground truth labels $q(\mathbf{g}) \in \{0,1\}$ of grasp success/failure and a binary cross-entropy loss
\begin{equation}
\label{eq:qual}
	\mathcal{L}_{qual} = \mathcal{L}_{CE}\left(\hat{q}({\mathbf{g}}), q({\mathbf{g}})\right)
\end{equation}

\noindent\textbf{Local supervision using ground truth surfaces.}~ Learning grasping functions based on surface features poses a challenge. Since we rely on local neural surface point rendering to pick appropriate point-wise features, the scene reconstruction needs to be accurate. A straightforward approach would be to train the grasp quality network $h^{q}_{\omega}$ in a subsequent step after the convergence of the scene reconstruction network $f^{sg}_{\theta}$. However, we wish to train grasp quality and scene reconstruction \textit{jointly} to ensure that the implicit geometric features $\psi$ also capture information relevant for assessing grasp quality. Thus, we propose to use additional local surface supervision at training time. During the data generation in simulation, we obtain the ground truth surface points for each grasp: $\left\{\mathbf{p}^{\mathbf{g}}_{gt\ i}\right\}_{i=1}^{N}$. We then add noise to these points, as visualized in Fig~\ref{fig:surf}c. These noisy surface points $\left\{\mathbf{p}^{\mathbf{g}}_{noisy\ i}\right\}_{i=1}^{N}$ serve two purposes. First, we can \textit{train a robust grasp quality network} using these noisy surface points and their features $\psi$ using the loss from \Cref{eq:qual}. Thus we only need to perform the local neural surface rendering at \textit{inference time} while also ensuring regularization against imperfect neural surface renders. Second, we add additional supervision to the scene reconstruction using the occupancy values for these dense yet noisy surface points $o(\mathbf{p}_{noisy})$, refining the occupancy in scene parts that interact with the gripper during grasping (cf. Fig.~\ref{fig:surf}). We use a loss function $\mathcal{L}_{local}$, similar to \Cref{eq:loss_occ}, to additionally train \textit{local object-part reconstruction} from these noisy surface points.

\noindent\textbf{Grasp affordance prediction}  We also experiment with extending our approach to, additionally, implicitly predict grasp \textit{affordances} towards task-oriented grasping. Using the local rendered features, we train another decoder network $d_{\gamma}^{a}$ with parameters $\gamma$ to predict the grasp affordance $\hat{a}(\mathbf{g})$. The affordances belong to a set of $K$ object affordance classes from 3D AffordanceNet~\cite{deng20213d}. Given that a grasp can have more than one affordance, this becomes a multi-label classification problem for which we can use separate heads to predict the probability of each affordance class $\left\{a_k\right\}_{k=1}^K$. For this, we use another binary cross entropy loss between the predicted affordance $\hat{a}_k(\mathbf{g})= d_{\gamma}^{a_k}\bigl( \left\{\mathbf{p}^{\mathbf{g}}_{i}, \psi(\mathbf{x},\mathbf{p}^{\mathbf{g}}_{i})\right\}_{i=1}^{N} \bigr)$ and the ground truth label $a_k(\mathbf{g}) \in \{0,1\}$ for every class $k$. Moreover, since the multi-label dataset is imbalanced and can contain many more negative labels than positive ones, we add a dice loss as proposed in \cite{deng20213d}. This loss ensures that the total error in the prediction of positive and negative labels in a batch is weighted equally. The overall affordance loss is then
\begin{equation*}\label{eq:aff}
    \mathcal{L}_{aff} = \overset{K}{\underset{k=1}{\sum}} \bigl( \mathcal{L}_{CE}\left(\hat{a_k}({\mathbf{g}}), a_k({\mathbf{g}})\right) + \mathcal{L}_{DICE}\left(\hat{a_k}({\mathbf{g}}), a_k({\mathbf{g}})\right) \bigr) .
\end{equation*}
\color{black}

\begin{table*}[t]
\setlength{\tabcolsep}{2pt}
\notsotiny
\centering
\captionsetup{justification=centering}
  \caption{Comparative results of NeuGraspNet vs. baselines on Pile and Packed scenes~\cite{Breyer2020} (5 seeds)}  \label{baselines-table}
\begin{adjustbox}{width=\textwidth,center}
  \begin{tabular}{@{}lccc|cc|cc|cc|cc|cc@{}}
    \hline
     {} & {} & \multicolumn{6}{c}{Pile scenes} & \multicolumn{6}{|c}{Packed scenes}\\
    \hline
    {} & {} & \multicolumn{2}{c}{Fixed Top View}&\multicolumn{2}{c}{Random View}&\multicolumn{2}{c}{Hard View} & \multicolumn{2}{|c}{Fixed Top View} & \multicolumn{2}{c}{Random View} & \multicolumn{2}{c}{Hard View}\\
    \cline{3-14}
    Method & Type & GSR (\%) & DR (\%) & GSR (\%)& DR (\%)& GSR (\%)& DR (\%) & GSR (\%) & DR (\%) & GSR (\%)& DR (\%)& GSR (\%)& DR (\%) \\
    \hline
    NeuGraspNet (ours) & Implicit & \textbf{86.51 $\pm$ 1.42} & \textbf{83.52 $\pm$ 2.24} & \textbf{85.05 $\pm$ 1.25} & \textbf{84.37 $\pm$ 1.52} & \textbf{73.95 $\pm$ 1.26} & \textbf{70.67 $\pm$ 1.69} & \textbf{97.65 $\pm$ 0.92} & \textbf{93.16 $\pm$ 1.48} & \textbf{92.49 $\pm$ 1.41} & \textbf{91.74 $\pm$ 1.24} & 78.76 $\pm$ 1.89 & 82.80 $\pm$ 1.50\\
    \hline
    \hline       
    PointNetGPD \cite{Liang2019} & Implicit & 79.79 $\pm$ 2.28 & 77.81 $\pm$ 2.79 & 70.94 $\pm$ 3.12 & 68.88 $\pm$ 3.11 & 47.42 $\pm$ 3.40 & 36.02 $\pm$ 3.68 & 81.14 $\pm$ 2.52 & 86.04 $\pm$ 1.50 & 71.94 $\pm$ 1.38 & 76.23 $\pm$ 2.51 & 25.15 $\pm$ 2.61 & 14.18 $\pm$ 1.50\\
     Scene-render+PNGPD & Implicit & 79.64 $\pm$ 2.35 & 78.02 $\pm$ 1.37 & 71.85 $\pm$ 1.76 & 68.45 $\pm$ 3.74 & 52.72 $\pm$ 4.89 & 40.77 $\pm$ 4.10 & 80.16 $\pm$ 2.33 & 84.93 $\pm$ 2.40 & 73.23 $\pm$ 1.45 & 76.86 $\pm$ 2.11 & 39.10 $\pm$ 4.47 & 25.38 $\pm$ 2.61\\
    VGN~\cite{Breyer2020} & Semi-Implicit & 77.44 $\pm$ 2.15 & 63.98 $\pm$ 5.03 & 78.48 $\pm$ 1.45 & 74.09 $\pm$ 1.16 & 68.46 $\pm$ 2.55 & 64.14 $\pm$ 4.37 & 83.42 $\pm$ 0.85 & 54.08 $\pm$ 5.35 & 78.11 $\pm$ 0.81 & 60.13 $\pm$ 1.96 & 70.27 $\pm$ 3.18 & 38.57 $\pm$ 3.40 \\
    EdgeGraspNet \cite{Huang2023} & Implicit & 80.25 $\pm$ 1.41 & 83.18 $\pm$ 1.43 & 78.76 $\pm$ 1.16 & 80.89 $\pm$ 2.65 & 68.11 $\pm$ 2.63 & 69.32 $\pm$ 3.79 & 85.09 $\pm$ 2.48 & 85.36 $\pm$ 2.74 & 86.06 $\pm$ 0.75 & 86.51 $\pm$ 0.92 & \textbf{81.61 $\pm$ 1.41} & \textbf{85.11 $\pm$ 0.84} \\
    GIGA~\cite{Jiang2021}  & Semi-Implicit & 82.92 $\pm$ 2.08 & 73.58 $\pm$ 2.93 & 78.67 $\pm$ 1.86 & 75.99 $\pm$ 1.79 & 69.13 $\pm$ 4.43 & 64.83 $\pm$ 5.63 & 96.05 $\pm$ 0.20 & 76.81 $\pm$ 3.21 & 87.99 $\pm$ 0.84 & 75.64 $\pm$ 2.75 & 73.87 $\pm$ 1.57 & 68.52 $\pm$ 4.49

    \\
    \hline
  \end{tabular}
  \footnotetext[1]{Footnote}

  \end{adjustbox}
\end{table*}

\begin{table*}[t]
\setlength{\tabcolsep}{2pt}
\notsotiny
\centering
\captionsetup{justification=centering}
  \caption{Ablation results of NeuGraspNet on Pile and Packed scenes~\cite{Breyer2020} (5 seeds)}  \label{ablations-table}
\begin{adjustbox}{width=\textwidth,center}
  \begin{tabular}{@{}lcc|cc|cc|cc|cc|cc@{}}
    \hline
     {} & \multicolumn{6}{c}{Pile scenes} & \multicolumn{6}{|c}{Packed scenes}\\
    \hline
    {} & \multicolumn{2}{c}{Fixed Top View}&\multicolumn{2}{c}{Random View}&\multicolumn{2}{c}{Hard View} & \multicolumn{2}{|c}{Fixed Top View} & \multicolumn{2}{c}{Random View} & \multicolumn{2}{c}{Hard View}\\
    \cline{2-13}
    Method & GSR (\%) & DR (\%) & GSR (\%)& DR (\%)& GSR (\%)& DR (\%) & GSR (\%) & DR (\%) & GSR (\%)& DR (\%)& GSR (\%)& DR (\%) \\
    \hline
    NeuGraspNet (ours) & 86.51 $\pm$ 1.42 & 83.52 $\pm$ 2.24 & \textbf{85.05 $\pm$ 1.25} & \textbf{84.37 $\pm$ 1.52} & \textbf{73.95 $\pm$ 1.26} & \textbf{70.67 $\pm$ 1.69} & \textbf{97.65 $\pm$ 0.92} & \textbf{93.16 $\pm$ 1.48} & \textbf{92.49 $\pm$ 1.41} & \textbf{91.74 $\pm$ 1.24} & \textbf{78.76 $\pm$ 1.89} & \textbf{82.80 $\pm$ 1.50} \\
\hline
\hline
    No-scene-render & 85.79 $\pm$ 1.38 & 83.44 $\pm$ 2.40 & 83.57 $\pm$ 1.91 & 83.06 $\pm$ 1.50 & 69.71 $\pm$ 2.48 & 65.61 $\pm$ 3.37 & 97.18 $\pm$ 1.14 & 92.80 $\pm$ 1.75 & 89.83 $\pm$ 1.79 & 90.19 $\pm$ 1.57 & 56.43 $\pm$ 4.62 & 29.08 $\pm$ 3.10 \\
    
    No-local-render & 79.83 $\pm$ 2.06 & 77.22 $\pm$ 2.72 & 77.04 $\pm$ 2.57 & 76.17 $\pm$ 2.51 & 63.51 $\pm$ 3.05 & 58.24 $\pm$ 3.14 & 96.31 $\pm$ 0.93 & 92.17 $\pm$ 1.43 & 89.81 $\pm$ 1.37 & 90.10 $\pm$ 0.70 & 73.86 $\pm$ 3.31 & 76.12 $\pm$ 2.16 \\

    No-shared-features & 80.72 $\pm$ 1.96 & 78.15 $\pm$ 2.672 & 77.20 $\pm$ 2.36 & 78.01 $\pm$ 2.47 & 64.44 $\pm$ 2.96 & 59.34 $\pm$ 3.17 & 95.44 $\pm$ 0.96 & 90.26 $\pm$ 1.48 & 88.71 $\pm$ 1.43 & 89.11 $\pm$ 0.83 & 71.24 $\pm$ 3.51 & 69.06 $\pm$ 2.33 \\
    
    No-local-occ & \textbf{86.62 $\pm$ 1.75} & \textbf{83.74 $\pm$ 2.41} & 84.37 $\pm$ 1.51 & 83.72 $\pm$ 0.89 & 72.08 $\pm$ 1.47 & 69.10 $\pm$ 2.13 & 96.53 $\pm$ 1.53 & 92.24 $\pm$ 1.37 & 90.62 $\pm$ 2.13 & 90.29 $\pm$ 1.56 & 78.14 $\pm$ 2.04 & 80.70 $\pm$ 1.73 \\
    
    No rendering & 73.59 $\pm$ 1.58 & 72.92 $\pm$ 2.82 & 73.36 $\pm$ 0.84 & 72.79 $\pm$ 1.38 & 56.52 $\pm$ 1.11 & 
    50.30 $\pm$ 2.75 & 93.65 $\pm$ 0.76 & 91.46 $\pm$ 0.37 & 87.32 $\pm$ 1.89 & 88.34 $\pm$ 1.47 & 52.17 $\pm$ 4.12 & 26.88 $\pm$ 2.31\\
    
    


    \hline
    \vspace{-0.5cm}
  \end{tabular}
  \end{adjustbox}
\end{table*}

\subsection{Implementation}
Our network uses a convolutional occupancy network encoder that encodes a scene Truncated Signed Distance Field (TSDF) (processed from the input pointcloud) into an implicit 3D feature space. For scene reconstruction, the decoder $f^{sg}_{\theta}$ is a ResNet-based fully-connected network (as in~\cite{Peng2020}). For grasp quality, the decoder $h_{\omega}^{q}$ uses a point network ~\cite{Qi2017}.
\\
The overall loss during training includes the occupancy loss $\mathcal{L}_{occ}$, quality loss $\mathcal{L}_{qual}$ as well as the additional local occupancy loss $\mathcal{L}_{local}$, each weighted by factors $w_i$,
\begin{equation}\label{eq:total_loss}
    \mathcal{L}_{NeuGrasp} = w_{occ}\mathcal{L}_{occ} + w_{qual}\mathcal{L}_{qual} + w_{local}\mathcal{L}_{local}.
\end{equation}
For affordance prediction, we also use the affordance loss to train separate heads for each affordance class:
\begin{equation}\label{eq:total_loss_a}
	\mathcal{L}_{NeuGraspAff} = \mathcal{L}_{NeuGrasp} + w_{aff}\mathcal{L}_{aff}.
\end{equation}

Additional implementation details are provided in Appendix A. 
We also experiment with incorporating rotational SO(3) equivariance in the encoder~\cite{Deng2021} and using different point network decoder architectures such as DGCNN~\cite{Wang2019}. Experiments with these architectures are provided in Appendix C.
\section{Experiments}

While evaluating our proposed approach, we aim to answer the following questions:
\begin{enumerate}
    \item Can NeuGraspNet outperform representative implicit and non-implicit baseline 6DoF grasping methods?
    \item What are the contributions of our rendering procedure at scene-level and at the local-level towards predicting accurate grasp quality?
    \item Given that our method utilizes scene reconstruction, how well does the method \textit{generalize} to unseen scenes with hard test objects where precise reconstruction is challenging?
    \item How well does our approach extend to predicting grasp affordances toward task-oriented grasping?
    \item How well does our method, trained in simulation, transfer to the real world?
\end{enumerate}

\color{black}
\subsection{Grasp Quality Prediction}\label{exp:sim}
\noindent\textbf{Experimental setup.}~ For training and evaluating NeuGraspNet, we first use the popular simulation benchmark from VGN~\cite{Breyer2020}, also used in \cite{Jiang2021, Huang2023, chisari2023centergrasp}, that uses the PyBullet simulator~\cite{Coumans2021}. 
In the VGN setup, simulated objects are spawned in a random `pile' or are `packed' upright on a surface. The benchmark consists of 343 household objects split into 303 training and 40 testing objects. On average, five objects are spawned per scene. The only modification we do to make the task more realistic and challenging is to \textit{randomize} the camera viewing angle when generating the scene TSDF or pointcloud. In addition to the VGN household object set, we also evaluate our method and baselines on test objects from the Evolved Grasping Analysis Dataset (EGAD)~\cite{morrison2020egad} \textit{without re-training the networks}. EGAD contains objects of varying shape complexities, and testing on them helps determine the generalizability of our method, especially to situations where precise reconstruction is challenging.\color{black}

\noindent\textbf{Data generation and training.}~ Since our network is implicit in all 6 grasp dimensions (unlike semi-implicit methods~\cite{Breyer2020,Jiang2021}), we create a dataset with a larger number of grasps per scene to learn to discriminate grasps in SE(3). We execute grasp candidates sampled using the method from~\Cref{subsec:gpg} on simulated scenes and train both NeuGraspNet and the baselines on this dataset.
We generate a dataset of 1.4 million grasps in 33,313 scenes for `pile' scenes and 1.2 million grasps in 33,534 scenes for `packed' scenes, balanced with both successful and unsuccessful grasps. We also sample 100,000 occupancy points per scene to train the scene reconstruction. Further details about the dataset generation are provided in Appendix B.

\noindent\textbf{Baselines \& metrics.}~ We compare our method against strong baselines that use various representations for predicting grasp quality. Due to the scope of our work for fully implicit grasp detection, we emphasize comparison with implicit or semi-implicit methods. We compare against \textbf{PointNetGPD}~\cite{Liang2019}, which operates directly on pointclouds, and experiment with adding our scene completion pipeline as preprocessing for PointNetGPD, termed \textbf{Scene-render+PNGPD}. We also compare against the current state-of-the-art \textbf{EdgeGraspNet}~\cite{Huang2023} that proposes a contact-edge-based grasp representation. Note that for EdgeGraspNet, we generate the dataset as per the sampling strategy in~\cite{Huang2023} as it requires grasp samples on edges.
Additionally, we compare against the semi-implicit methods \textbf{VGN}~\cite{Breyer2020} and \textbf{GIGA}~\cite{Jiang2021}, with the latter considering scene reconstruction as an auxiliary loss. We train VGN and GIGA with high-resolution TSDFs (as also done for NeuGraspNet), as it leads to better performance. We also run ablations of our model with different settings, e.g., with and without scene-level rendering, etc., to demonstrate the efficacy of our full model. We train all methods from scratch on our dataset with random-view (random elevation between 15 and 75 degrees) pointcloud or TSDF inputs for a fair comparison. Though our work focuses on implicit grasp quality prediction, for completeness, we also compare against two popular generative grasping networks, \textbf{AnyGrasp}~\cite{fang2022anygrasp} and \textbf{ContactGraspNet}~\cite{Sundermeyer2021}. For these, we use pre-trained networks and hence mainly test on the EGAD object setting where all the networks are evaluated without re-training. Further details about the training and evaluation setup are provided in Appendix B.

The main focus of this work is to evaluate the representation capability of NeuGraspNet, which learns the interaction between scene/object geometry and the robot end-effector for detecting high-quality 6DoF grasps. Our benchmarking setup consists, therefore, of three settings. (I) \textbf{Fixed Top View}: In this setting, the camera is placed at a fixed viewing angle: a 60-degree elevation, and at a distance of twice the workspace width (width in this setup is 30cm), as in~\cite{Jiang2021}. (II) \textbf{Random view}: This setup resembles the one of training; the viewing elevation is sampled between 15 and 75-degrees, and the distance of the camera is between 1.6 and 2.4 times the workspace width. This setting is harder than the one used in~\cite{Huang2023} that considers angles between 30 and 45 degrees. (III) \textbf{Hard view}: While we sample the camera's distance like the random view setting, we fix the viewing angle at 15 degrees. We argue that such views are realistic for robots like mobile manipulators that can move in the scene to grasp scattered objects with a limited field of view. Other methods that use fixed manipulators operating on tabletop surfaces have not considered such a setting. All these settings are tested on the `pile' and `packed' scenes of~\cite{Breyer2020} and `pile' scenes with EGAD~\cite{morrison2020egad} objects. To reconstruct the scene, we use N=6 virtual cameras in a circular trajectory around the scene. From our experiments, we note that using less than 4 virtual cameras is insufficient to reconstruct the surfaces well for the considered grasping scenes, while more than 6 virtual cameras do not provide any benefit. We use the same grasping metrics as~\cite{Jiang2021,Breyer2020,Huang2023}, namely, the Grasp Success Rate (GSR), i.e., percentage of successful grasp executions w.r.t. the total attempts, and the Declutter Rate (DR), i.e., percentage of objects successfully removed to the number of total objects presented in the scenes. We perform 100 simulation rounds for 5 random seeds in all cases and report the average percentage and standard deviation per metric. As per the experiment setup of~\cite{Breyer2020}, a simulation round ends when (a) no objects remain, (b) no high-quality grasps are found, or (c) two consecutive failed grasp attempts have occurred.

\begin{table*}[ht!]
\scriptsize
\centering
\captionsetup{justification=centering}
  \caption{Grasping performance in simulated Pile scenes with objects from the EGAD~\cite{morrison2020egad} dataset. The networks are not re-trained but directly tested on EGAD objects of different complexities (5 seeds)} \label{egad-table}
  \begin{tabular}{@{}lccc|cc|cc@{}}
    \hline
    {} & {} & \multicolumn{2}{c}{Fixed Top View}&\multicolumn{2}{c}{Random View}&\multicolumn{2}{c}{Hard View}\\
    \cline{3-8}
    Method & Type & GSR (\%) & DR (\%) & GSR (\%)& DR (\%)& GSR (\%)& DR (\%) \\
    \hline
    NeuGraspNet (ours) & Implicit & \textbf{79.76 $\pm$ 1.54} & \textbf{81.82 $\pm$ 1.45} & \textbf{75.24 $\pm$ 1.01} & \textbf{76.80 $\pm$ 1.29} & \textbf{67.21 $\pm$ 2.86} & \textbf{62.04 $\pm$ 4.94} \\
\hline
\hline

    PointNetGPD \cite{Liang2019} & Implicit & 65.71 $\pm$ 3.20 & 64.73 $\pm$ 3.19 & 59.05 $\pm$ 2.50 & 56.19 $\pm$ 4.68 & 37.64 $\pm$ 1.16 & 25.99 $\pm$ 0.95 \\

    VGN \cite{Breyer2020} & Semi-Implicit & 13.59 $\pm$ 1.16 & 11.71 $\pm$ 0.59 & 11.42 $\pm$ 0.54 & 9.98 $\pm$ 0.25 & 5.56 $\pm$ 0.56 & 4.72 $\pm$ 0.72 \\

    EdgeGraspNet \cite{Huang2023} & Implicit & 40.59 $\pm$ 2.38 & 31.99 $\pm$ 2.37 & 40.75 $\pm$ 1.97 & 31.81 $\pm$ 2.20 & 39.50 $\pm$ 1.91 & 30.03 $\pm$ 2.66 \\
        
    GIGA \cite{Jiang2021} & Semi-Implicit & 27.49 $\pm$ 3.05 & 17.65 $\pm$ 2.42 & 25.53 $\pm$ 2.84 & 16.03 $\pm$ 2.52 & 17.10 $\pm$ 1.96 & 9.28 $\pm$ 1.42 \\
    
    ContactGraspNet \cite{Sundermeyer2021} &Generative &48.98 $\pm$ 4.27 &40.85 $\pm$ 2.96 &42.25 $\pm$ 2.35 &38.66 $\pm$ 2.48 &30.46 $\pm$ 2.74 &22.64 $\pm$ 1.28 \\
    
    AnyGrasp \cite{fang2022anygrasp} & Generative & 68.46 $\pm$ 2.50 & 68.06 $\pm$ 2.74 & 66.09 $\pm$ 1.01 & 66.83 $\pm$ 1.86 & 56.23 $\pm$ 0.97 & 47.43 $\pm$ 0.95 \\

     


    \hline
  \end{tabular}
\end{table*}

\begin{figure*}[h]
     \centering
     \vspace{-0.385cm}
     \begin{subfigure}[b]{0.175\textwidth}
         \centering
         \includegraphics[width=\textwidth,trim={2cm 2cm 2cm 1cm},clip] {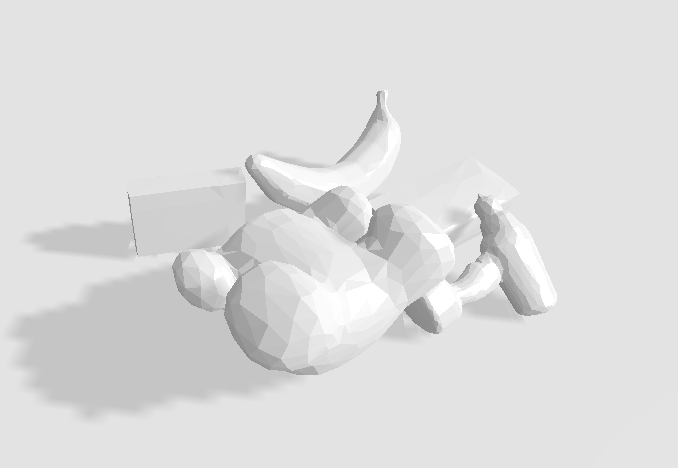}
        \vspace{-0.4cm}
     \end{subfigure}
     \begin{subfigure}[b]{0.175\textwidth}
         \centering
         \includegraphics[width=\textwidth,trim={3cm 9cm 1cm 3cm},clip]{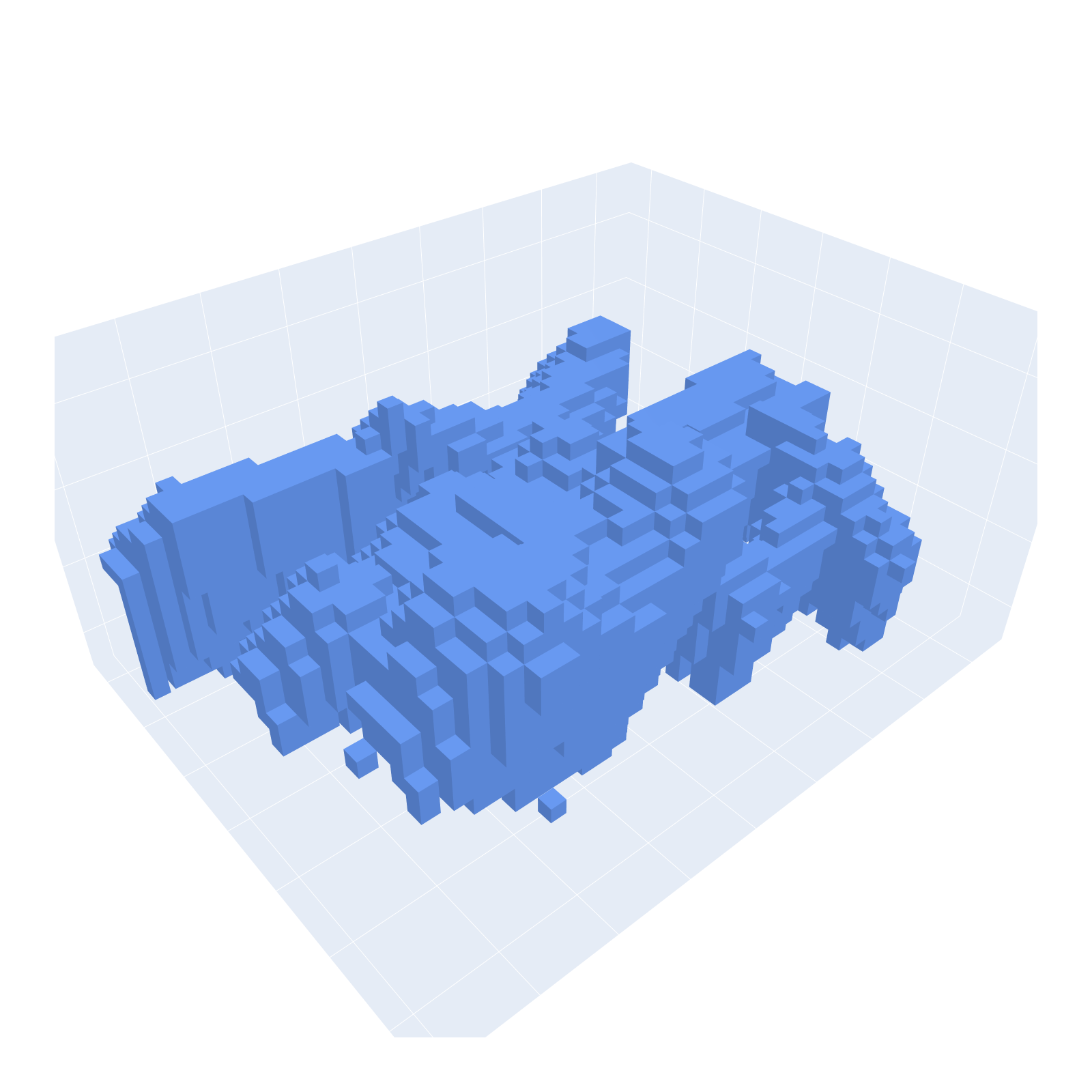}
        \vspace{-0.4cm}
     \end{subfigure}
      \begin{subfigure}[b]{0.175\textwidth}
         \centering
         \includegraphics[width=\textwidth,trim={0cm 6cm 3cm 0cm},clip]{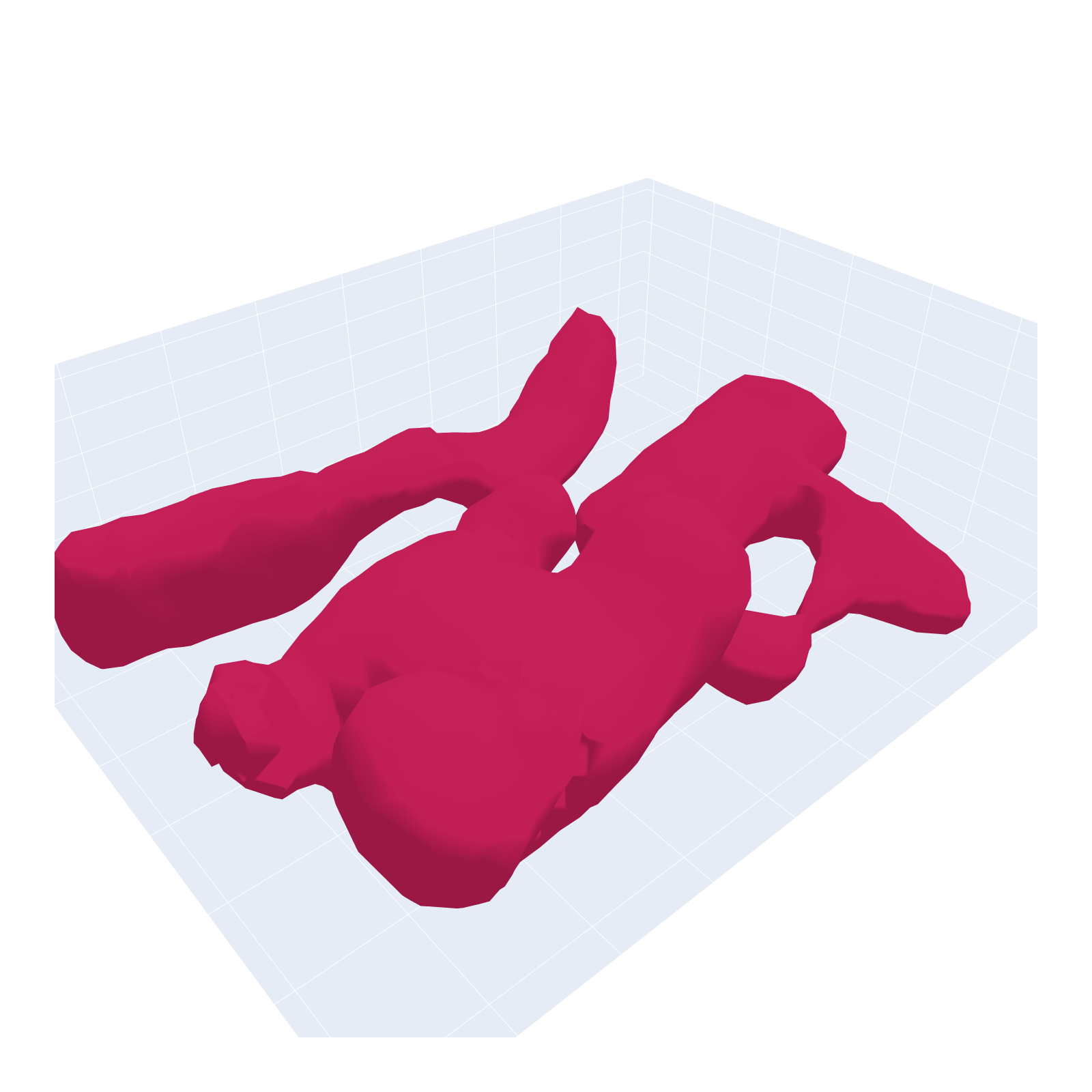}
        \vspace{-0.4cm}
     \end{subfigure}
     \begin{subfigure}[b]{0.175\textwidth}
         \centering
         \includegraphics[width=\textwidth,trim={5cm 4cm 5cm 0cm},clip]{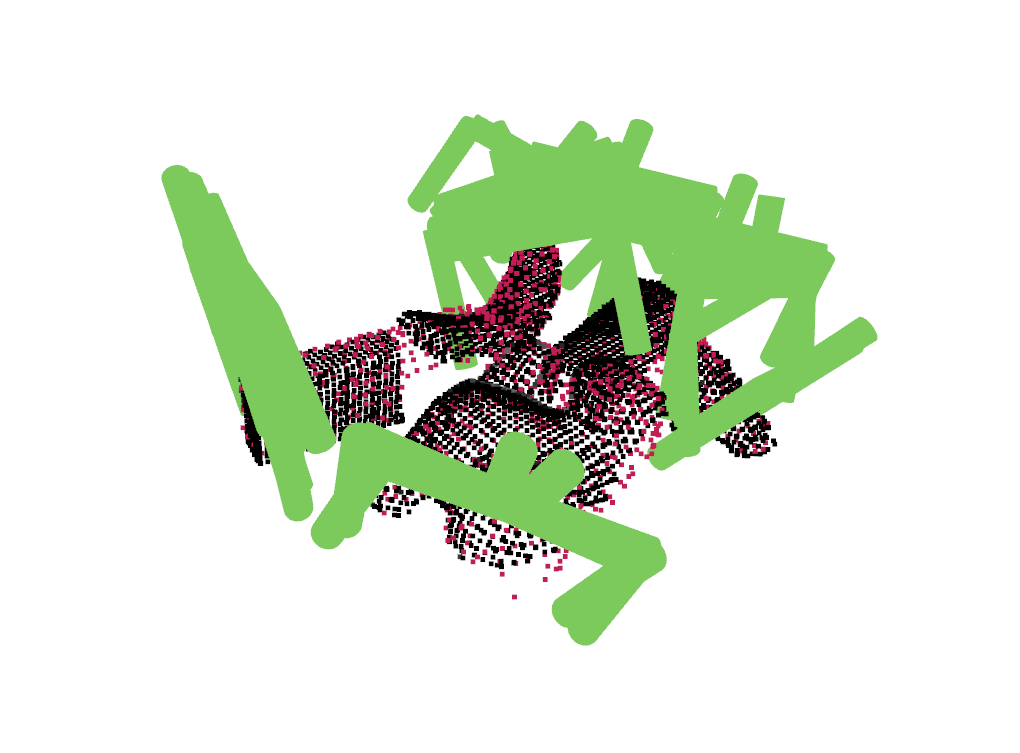}
        \vspace{-0.4cm}
     \end{subfigure}
      \vspace{-0.2cm}
      
      \begin{subfigure}[b]{0.175\textwidth}
         \centering
         \includegraphics[width=\textwidth,trim={2cm 1cm 2cm 1cm},clip] {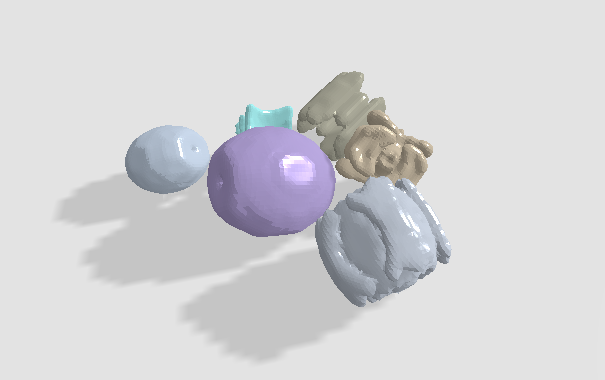}
        \vspace{-0.4cm}
         \caption{Scene}
     \end{subfigure}
     \begin{subfigure}[b]{0.175\textwidth}
         \centering
         \includegraphics[width=\textwidth,trim={1cm 10cm 1cm 0cm},clip]{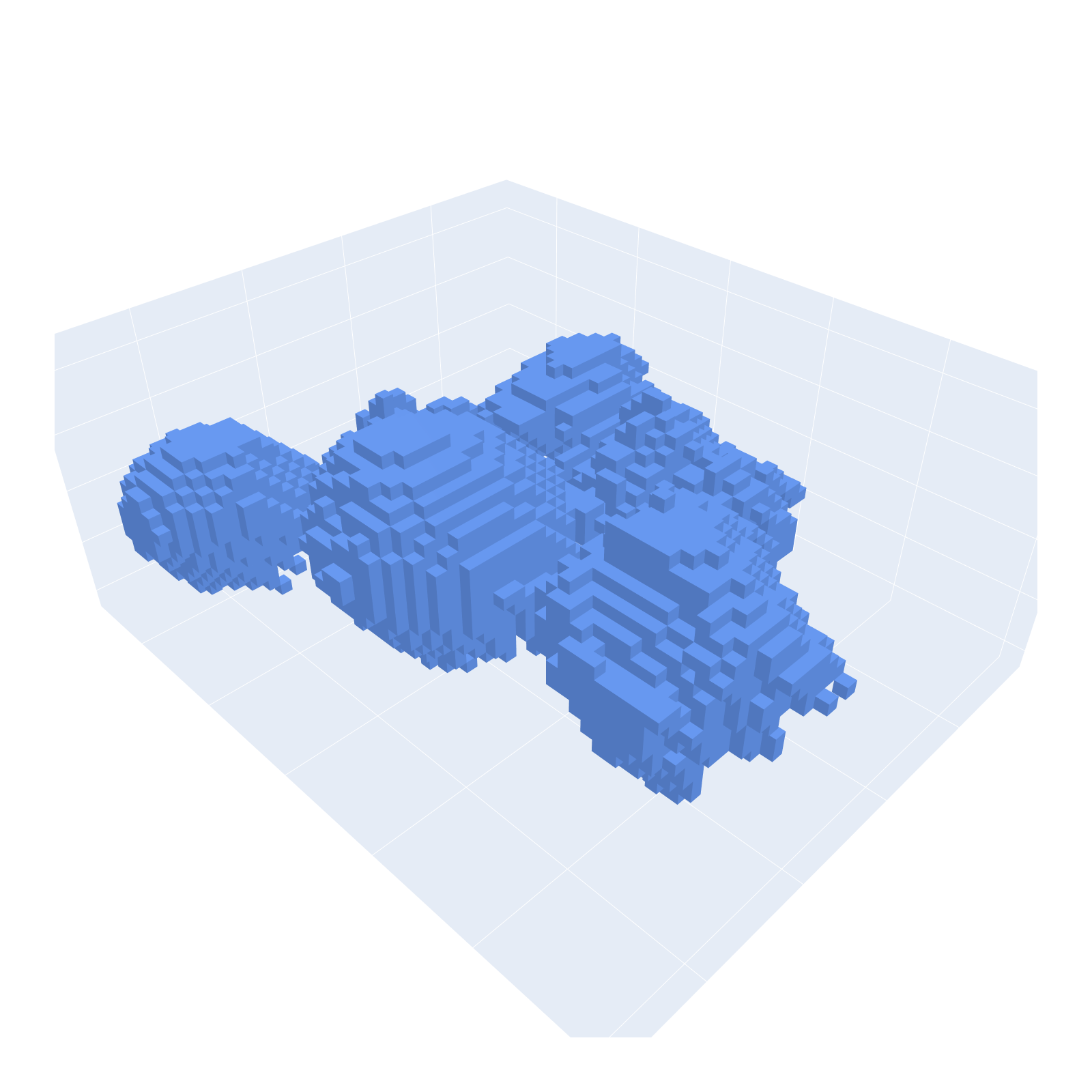}
        \vspace{-0.4cm}
         \caption{Input TSDF}
     \end{subfigure}
      \begin{subfigure}[b]{0.175\textwidth}
         \centering
         \includegraphics[width=\textwidth,trim={0cm 9cm 3cm 0cm},clip]{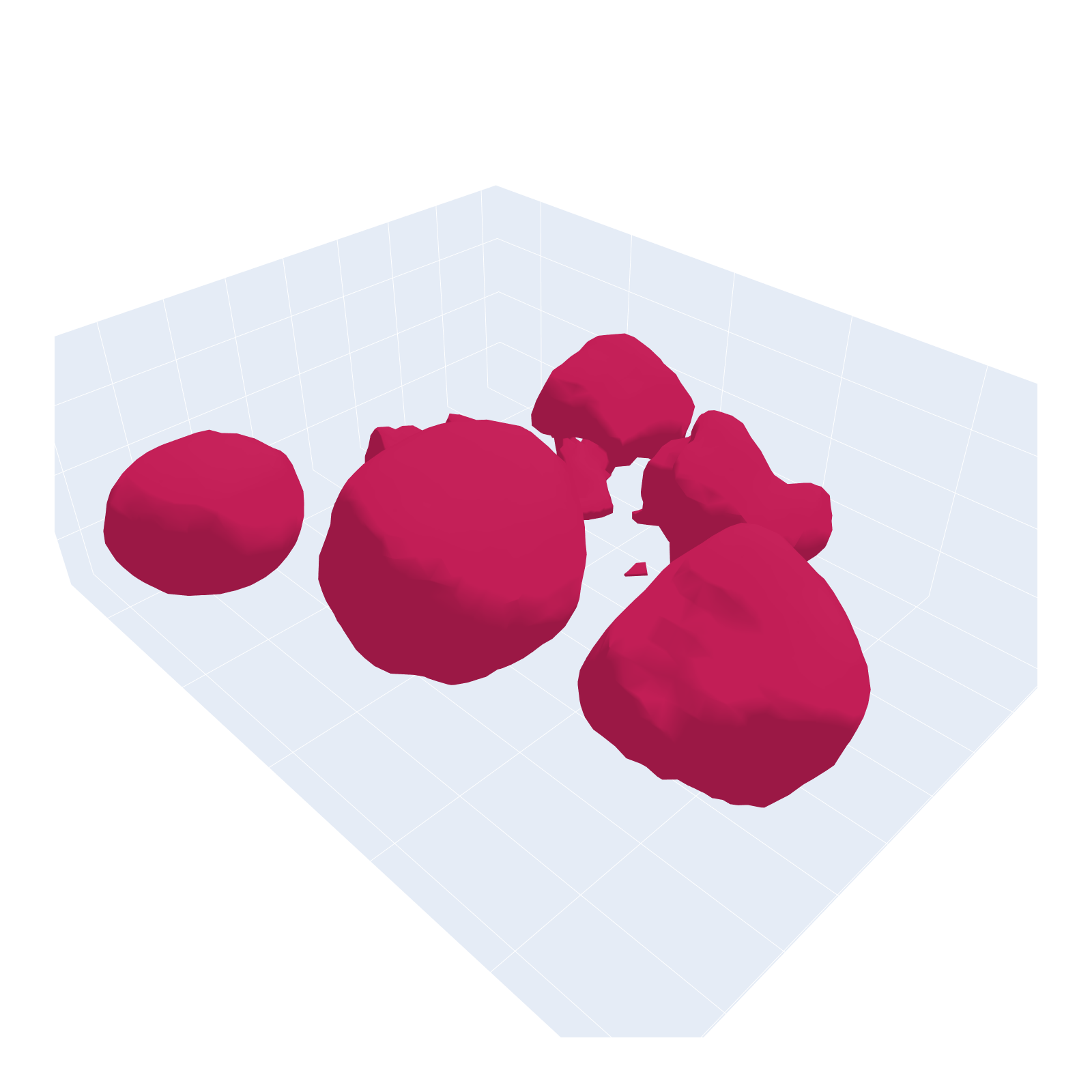}
        \vspace{-0.4cm}
         \caption{Reconstruction}
     \end{subfigure}
       \begin{subfigure}[b]{0.175\textwidth}
         \centering
         \includegraphics[width=\textwidth,trim={5cm 1cm 5cm 0cm},clip]{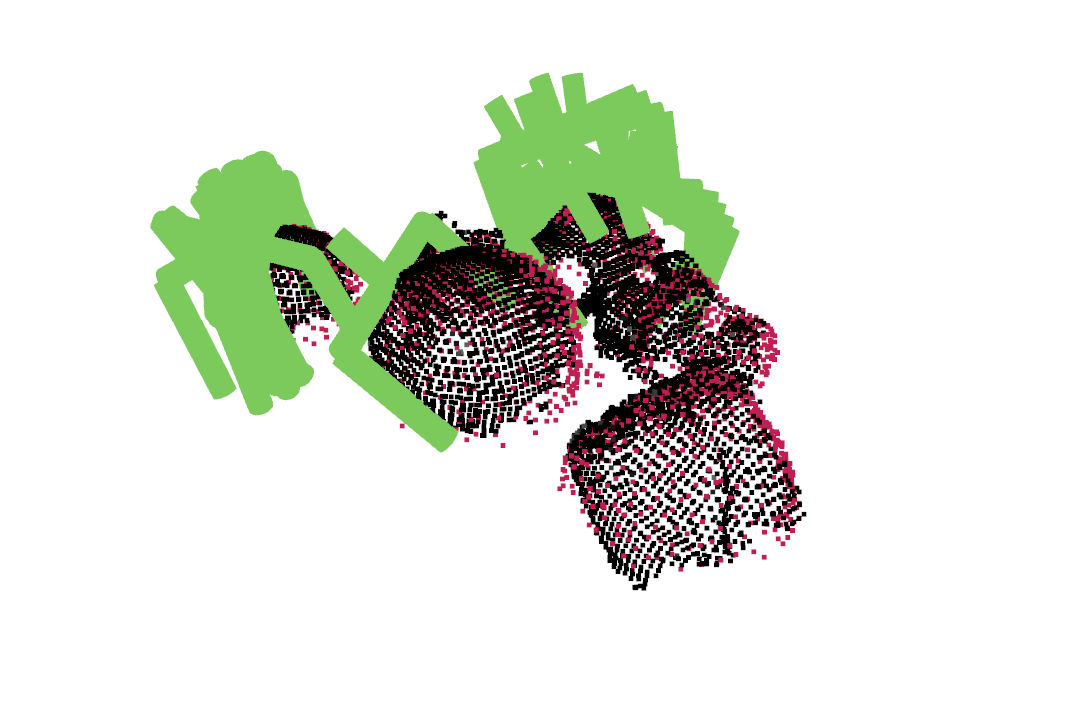}
        \vspace{-0.4cm}
         \caption{Detected Grasps}
     \end{subfigure}
    \caption{Example scene reconstructions \& detected grasps for unseen test objects from the VGN~\cite{Breyer2020} (top) and the EGAD~\cite{morrison2020egad} (bottom) datasets. We see that our network can sometimes create artifacts or is unable to reconstruct very fine details, especially for the hard EGAD objects. Nevertheless, even in these hard cases, our network is able to reconstruct the broad structure of the scene \& objects which results in the detection of good grasps (d).}
    \label{fig:scene-recons}
    \vspace{-0.3cm}
\end{figure*}

\subsubsection{Comparison with baselines on VGN~\cite{Breyer2020} scenes}
Observing Table~\ref{baselines-table}, in `pile' scenes, which are more unstructured, NeuGraspNet outperforms the baselines in all settings. EdgeGraspNet and GIGA perform similarly in GSR, but EdgeGraspNet has a higher DR, indicating fewer consecutive failures. In `packed' scenes, we observe a similar high performance by NeuGraspNet. However, there are interesting remarks regarding the baselines. GIGA still performs well in fixed top view settings, where most of the scene is still observed. VGN performs close to GIGA in random and hard views. PointNetGPD performs reasonably in the fixed top view setting since it uses the GPG~\cite{Gualtieri2016} grasp sampling strategy, which favors top grasps. However, it performs poorly in hard view settings since top surfaces of packed objects are rarely observed. Our version of PointNetGPD that uses scene completion, i.e. Scene-render+PNGPD, performs very similar to PointNetGPD in the fixed view cases, but shows improvements of up to 5-10\% in hard views. EdgeGraspNet showcases good performance in `packed' scenes, and with `packed' hard views, it outperforms NeuGraspNet by $\mathtt{\sim}2.5$\% in GSR and DR. Note that we test EdgeGraspNet with their proposed edge grasp sampling strategy. In `packed' scenes, this edge strategy favors side grasps, unlike GPG, which favors top grasps. This characteristic of GPG also hurts performance of NeuGraspNet in `packed' hard-view settings. While we have shown gains due to our scene-level GPG approach (cf. \Cref{subsec:gpg}), an improved geometrically-informed sampling strategy could lead to further improvements. Nevertheless, we demonstrate that our local point and feature set successfully encodes the gripper-object complementarity and outperforms all baselines in 5 out of 6 settings.

\subsubsection{Ablation study}
Table~\ref{ablations-table} presents an ablation of different components of NeuGraspNet for evaluating the contribution of each component of the network. As the results show, a model without any rendering (global or local), shown at the bottom of the table, performs the worst. In this case the grasp sampler relies only on the input pointcloud, and the grasp quality network struggles to discriminate in the SE(3) space. When only removing global scene-level rendering, we achieve good performance with fixed \& random views but see a significant drop with hard views. In hard views, much of the scene is unseen and the grasp sampler struggles to sample reasonable candidates. \textit{Removing the local rendering at the grasp level from NeuGraspNet hurts performance the most} (`No-local-render' in table), underlying the importance of local features that allow learning the interaction of object surface and gripper. Not using a shared feature space for the reconstruction and grasp quality prediction (`No-shared features') also drops performance. We also observe worse performance without local occupancy supervision, especially for DR in `packed' scenes, where objects are placed closely, and fine details around objects are needed. In this case, providing a refinement signal with local occupancy improves performance. Additional ablations are in Appendix C.

\begin{table*}[ht]
\scriptsize
\setlength{\tabcolsep}{3pt}
\centering
\captionsetup{justification=centering}
\centering
\captionsetup{justification=centering}
  \caption{Grasping \& affordance prediction performance in random-view scenes with PartNet objects~\cite{deng20213d} (5 seeds)} \label{affnet-table}
  \begin{tabular}{@{}lcc|cccccccc@{}}
    \hline
    Method & GSR (\%) & DR (\%) & \multicolumn{8}{c}{Grasp affordance prediction mAP (\%)} \\
     &  &  & Average & Handover & Cut & Stab & Lift & Wrap & Pour & Wear \\
    \hline
    NeuGraspAffNet (ours) & \textbf{71.78 $\pm$ 3.19} & \textbf{57.55 $\pm$ 3.79} & \textbf{64.21 $\pm$ 2.49} & 
    \textbf{84.34 $\pm$ 2.60} & 78.61 $\pm$ 5.89 & 78.58 $\pm$ 5.89 & \textbf{35.68 $\pm$ 7.96} & \textbf{31.70 $\pm$ 7.85} & \textbf{75.68 $\pm$ 8.44} & \textbf{64.91 $\pm$ 4.63}
    \\
    \hline
    \hline
    
    No-local-render (ours) & 57.48 $\pm$ 1.30 & 39.78 $\pm$ 1.81 & 53.45 $\pm$ 3.58 &
    82.86 $\pm$ 3.60 & \textbf{79.05 $\pm$ 5.91} & \textbf{79.01 $\pm$ 5.89} & 22.03 $\pm$ 8.44 & 21.71 $\pm$ 4.97 & 65.15 $\pm$ 5.96 & 24.35 $\pm$ 8.42
    \\

    

    VGN+Aff\cite{Breyer2020} & 16.11 $\pm$ 1.71 & 5.80 $\pm$ 0.87 & 28.73 $\pm$ 1.42 &
      56.92 $\pm$ 1.18 &
      19.06 $\pm$ 2.43 &
      19.06 $\pm$ 2.43 &
      21.10 $\pm$ 0.63 &
      27.33 $\pm$ 2.22 &
      13.23 $\pm$ 1.31 &
      44.38 $\pm$ 2.06    
    \\
      
    GIGA+Aff \cite{Jiang2021} & 53.72 $\pm$ 3.01 & 44.78 $\pm$ 2.65 & 42.64 $\pm$ 1.64 &
      83.23 $\pm$ 0.87 &
      29.26 $\pm$ 4.54 &
      29.27 $\pm$ 4.54 &
      34.39 $\pm$ 5.49 &
      30.92 $\pm$ 4.56 &
      34.71 $\pm$ 6.67 &
      56.67 $\pm$ 5.32
    \\

    
    \hline
  \end{tabular}
  \vspace{-0.3cm}
  \end{table*}

  \begin{table}[h!]
\caption{Grasping performance in real-world scenes over 12 scene decluttering rounds}
\label{table:real}
\centering
\begin{adjustbox}{width=0.99\columnwidth,center}
  \begin{tabular}{@{}lcc|cc@{}}
    \hline
    {} & \multicolumn{2}{c}{Pile scenes (Random View)} & \multicolumn{2}{|c}{Packed scenes (Random View)} \\
    \cline{2-5}
    Method & GSR (\%) & DR (\%) & GSR (\%) & DR (\%) \\
    \hline
    NeuGraspNet (ours) & \textbf{83.63} (46/55) & \textbf{76.36} (46/60) & \textbf{87.72} (50/57) & \textbf{90.90} (50/55)\\
\hline
\hline
    EdgeGraspNet \cite{Huang2023} & 64.81 (35/54) & 58.33 (35/60) & 70.58 (36/51) & 65.45 (36/55) \\
    \hline
    
  \end{tabular}
  \end{adjustbox}
\vspace{-0.5cm}
\end{table}

\subsubsection{Comparison with baselines on EGAD~\cite{morrison2020egad}}
\Cref{egad-table} presents evaluations on piled EGAD~\cite{morrison2020egad} object scenes without re-training the networks. We note a very significant performance drop in all methods due to the diversity in complexity of EGAD objects. However, NeuGraspNet sees the smallest drop, showing greater robustness than baselines. Example visualizations of reconstructions and detected grasps are provided in~\Cref{fig:scene-recons}. We note that even in difficult reconstruction scenarios, NeuGraspNet recovers the broad scene structure, leading to good grasps being detected. The performance of implicit and semi-implicit baselines is similar to the `pile' scenes with VGN objects. The generative AnyGrasp performs better than most other baselines but still significantly lags behind NeuGraspNet in both GSR and DR while ContactGraspNet performs poorly in hard views but comparably to EdgeGraspNet in fixed and random view settings.

\subsection{Grasp Affordance Prediction}
\label{subsec:aff_exp}

We evaluate the extension of our method to grasp affordance prediction in depth pointclouds and provide initial results.

\noindent\textbf{Data generation and training.} There exists a lack of a standard, reproducible benchmark for task-oriented grasping and grasp affordance prediction. Very few methods try to predict grasp affordances in 3D pointclouds while also operating in cluttered scenes. Thus, we generate a grasp affordance dataset using object and semantic affordances from 3DAffordanceNet~\cite{deng20213d}. We use the same simulation setup as above but use piled objects from~\cite{deng20213d}, i.e., PartNet objects~\cite{Mo_2019_CVPR}. For simulated successful grasps of an object, we assign grasp affordances by checking grasp proximity to labeled affordance points from 3DAffordanceNet. Note that the dataset is multi-label, i.e., each grasp can have multiple assigned affordances. During data generation, we use a subset of object categories in 3DAffordanceNet that can be expected to be grasped, giving us 1,248 objects from categories `Knife', `Scissors', `Mug', `Bottle', `Bag', `Earphone', and `Hat'. While this reduces our study to these categories, we note that significant diversity still exists within the objects to determine the extendability of our approach to task-oriented grasping. We end up with 7 affordance classes for grasps on our chosen object set, namely `Handover', `Cut', `Stab', `Lift', `Wrap', `Pour', and `Wear'. We also remap affordances to be robot-centric, i.e., affordance labels denote actions that can be performed \textit{after} grasping the object. Our dataset contains 2.6 million grasps in 37,696 scenes. We note that PartNet objects are significantly harder to grasp since many of them are too small or large in their original scale. We train our network, called NeuGraspAffNet, and baselines with binary affordance classification heads for each class, given the multi-label nature of the data.
\\
\noindent\textbf{Metrics \& comparison with baselines.}
We split the objects in the dataset into 1096 train and 152 test objects and evaluate grasping success and accurate affordance prediction on scenes with test objects (100 simulation rounds for 5 random seeds). We report the GSR and DR and also include the mean Average Precision (mAP) for affordance classification per class. We also report the average mAP over affordance classes, i.e., the macro-average. 
Since few methods exist that perform scene-level and shape-based grasp affordance prediction, we compare against modified versions of VGN and GIGA with additional affordance heads \textbf{GIGA+Aff} and \textbf{VGN+Aff}. For completeness, we also compare against our ablation with no-local-rendering to validate if lack of local rendering has the same effect as in the grasp quality evaluations. The results are listed in~\Cref{affnet-table}. Consistent with the previous results, NeuGraspAffNet significantly outperforms the baselines regarding success and declutter rates. It also performs better in terms of affordance prediction with an average mAP of 64.21\%. Among the affordance classes, the `Wrap' class is the most difficult to accurately predict, while the `Handover' and `Cut' classes are the easiest. An example visualization of grasp affordance prediction is provided in Appendix C in~\Cref{fig:real_aff}. While we've shown the benefit of our approach compared to baselines, there still exists significant scope for improvement in task-oriented grasping for our random-view, scene-level setting. Using visual affordances from 2D RGB images, projecting them into 3D, and integrating them into our reconstruction pipeline is a promising direction for further research.

\color{black}
\begin{figure*}[h]
     \centering
     \vspace{-0.385cm}
     \hspace{-1cm}
     \begin{subfigure}[b]{0.54\textwidth}
         \includegraphics[width=\textwidth] {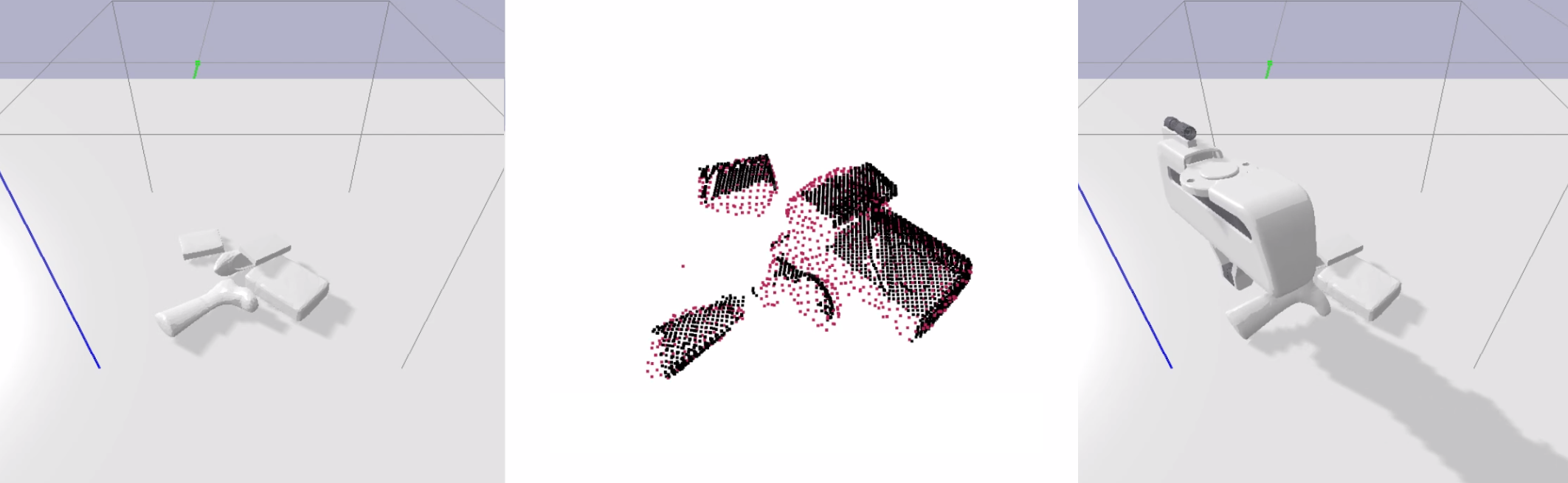}
         \caption{In this example with input depth from a challenging view (black pointcloud in the middle image), the scene reconstruction (red pointcloud) is not sufficient to avoid grasp failure (right image).}
     \end{subfigure}
     \hspace{0.1cm}
     \begin{subfigure}[b]{0.45\textwidth}
         \includegraphics[width=\textwidth] {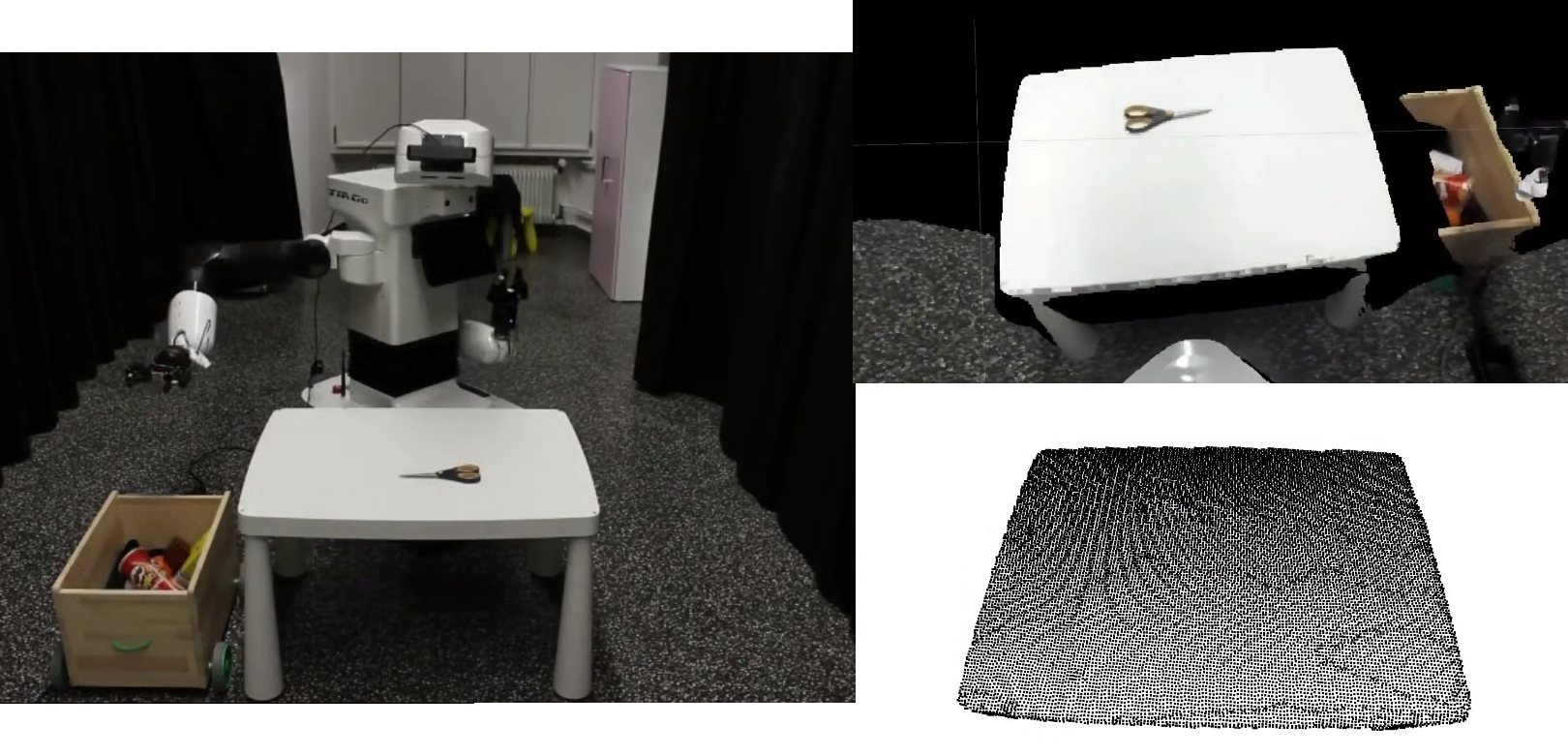}
         \caption{In this example, the thin scissor placed flat on the table is not perceived well by the depth camera mounted on the real robot head, leading to no grasps being detected.}
     \end{subfigure}
    
    \caption{Example failure cases observed in (a) simulated and (b) real-world experiments.}
    \label{fig:failures}
    \vspace{-0.3cm}
\end{figure*}

\subsection{Real-world Evaluation}
\label{subsec:real}
We perform real-world experiments with a mobile manipulator robot, TIAGo++, equipped with a head-mounted RGBD camera. The robot is placed in house-like scenarios with objects placed randomly in a  `pile' or `packed' setting. We use 16 objects from the YCB dataset \cite{calli2015benchmarking} for this experiment. For each setting, we perform 12 decluttering rounds and report the GSR and DR observed. \Cref{{fig:cover_tiago}} shows example grasp executions in random-view mobile manipulation scenarios such as grasping from a shelf or grasping from the top of a cupboard. We also compare to the best-performing baseline EdgeGraspNet~\cite{Huang2023}. \Cref{table:real} lists the real-world experiment results. We observe a GSR and DR of 83.63 \% and 76.36 \%, respectively, for `pile' scenes and a GSR and DR of 87.72 \% and 90.90 \%, respectively, for `packed' scenes. We also demonstrate our affordance prediction capabilities by grasping based on user-queried affordances. Demonstrations are provided in the supplementary material. 
We note a drop in performance compared to~\Cref{exp:sim} due to the following failure cases. The execution of grasps on objects of relatively large width can sometimes fail due to imprecise perception or execution by the mobile robot, hurting the GSR. On the other hand, very thin objects are not always clearly perceived by the random-view depth camera since it is difficult to differentiate between the object and the table, hurting the DR. Nevertheless, with the use of better camera depth estimation techniques, we expect to get even better performance on real-world de-cluttering of scenes. Additional details and results are provided in Appendix C and the project website: \href{https://sites.google.com/view/neugraspnet}{sites.google.com/view/neugraspnet}.

\begin{table}[t]
\scriptsize
\centering
\captionsetup{justification=centering}
\caption{Grasp sampling time and inference time comparison between methods (in milliseconds)}
\label{table:time}
  \begin{tabular}{@{}lccc@{}}
  
    \hline
    Method & Grasp sampling & Grasp quality inference & Total time \\
    \hline
    NeuGraspNet (ours) & 608 ms & 257 ms & 865 ms \\
\hline
\hline
    EdgeGraspNet \cite{Huang2023} & 401 ms & 30 ms & 431 ms \\

    No-scene-render (ours) & 161 ms & 257 ms & 418 ms \\

    ContactGraspNet \cite{Sundermeyer2021} &n.a. &280 ms &280 ms \\
    
    GIGA \cite{Jiang2021} & n.a. & 225 ms & 225 ms \\

    AnyGrasp \cite{fang2022anygrasp} & n.a. & 207 ms & 207 ms \\

    PointNetGPD \cite{Liang2019} & \textbf{170 ms} & \textbf{23 m}s & \textbf{193 ms} \\
    \hline
  \end{tabular}
  \vspace{-0.3cm}
\end{table}

\section{Conclusion and Limitations}
We presented NeuGraspNet, a novel, fully implicit 6DoF-grasp prediction method that re-interprets robotic grasping as surface rendering. NeuGraspNet predicts high-fidelity 6DoF grasps from any random single viewpoint of a scene. Our method exploits a learned implicit geometric scene representation to perform global and local surface rendering. This enables effective grasp candidate generation (using global features) and grasp quality prediction (using local features from a shared feature space). We demonstrated the superior performance of NeuGraspNet in simulated experiments of varying difficulty. We also explored the method's applicability to task-oriented grasping by predicting grasp affordances. Finally, we exhibited the real-world applicability of NeuGraspNet in mobile manipulator grasping experiments.
\\
Our method, however, comes with certain limitations. Firstly, as noted in section IV.C the network struggles to accurately perceive depth and reconstruct very thin objects. An example of this is provided in \Cref{fig:failures}. 
Moreover, NeuGraspNet requires a computational overhead due to the multiple surface rendering processes, reported in~\Cref{table:time}. Nevertheless, we posit that advances in neural rendering can make such methods increasingly accessible, even with onboard computing. Another limitation is that our model presumes a particular instantiation of gripper. While we have tested with two different grippers, the Franka Panda gripper and the Robotiq 2F-85, the model can struggle with a new gripper of a very different size. This problem could be mitigated by providing the gripper model or size as an input to the network instead, an addition that we wish to explore in future work. A further limitation that affects most grasp detection approaches is the open-loop nature of execution, without consideration of sudden scene changes. Thus, coupling grasp detection (perception) and manipulation (action) dynamically is part of our plans. Finally, while we show promising initial results with respect to shape-based grasp affordances, the reliability and diversity of predicted affordances can be further improved. Projecting pre-trained 2D RGB features, using recent approaches such as generalizable feature fields~\cite{ye2023featurenerf} into our pipeline is a promising direction toward any-scene task-oriented grasping. Moreover, the evaluation of downstream manipulation tasks after the grasp has succeeded is also an important metric to consider in our framework, one that we wish to explore further in future work. 

\section*{Acknowledgements}
This research received funding from the European Union's Horizon program under grant agreement no. 101120823, project MANiBOT and the German Research Foundation (DFG) Emmy Noether Programme (CH 2676/1-1). We also gratefully acknowledge the computing time provided to us on the high-performance computer Lichtenberg at the NHR Centers NHR4CES at TU Darmstadt, Germany. This is funded by the Federal Ministry of Education and Research, and the state governments participating based on the resolutions of the GWK for national high-performance computing at universities (\url{www.nhr-verein.de/unsere-partner}).

\newpage

\bibliographystyle{plainnat}
\bibliography{main_bib}

\clearpage
\appendices

\section{Network architecture and implementation details}
\label{appendix:a}

\subsection{Network architecture}\label{app:subsec:network}

Our network input is a pointcloud or Truncated Signed Distance Field (TSDF) captured from a single viewpoint. We use an encoder-decoder architecture with scene reconstruction and grasp quality prediction heads. We predict grasp quality based on features of the surface point set that the grasp interacts with. For this, we use a feature point network decoder to classify successful/unsuccessful grasps. The overall pipeline is in~\Cref{fig:pipeline}; our code will be open-sourced upon acceptance.

\subsubsection{Encoder}

We use a similar encoder architecture as Convolutional Occupancy Networks (ConvONets)~\cite{Peng2020}. If the input is a scene voxel grid (eg. TSDF), ConvONets use a simple 3D encoder with ResNets and U-Nets. If a pointcloud input is provided, a `local' PointNet is first used to encode the scene before the U-Net stage. We experiment with both versions of the encoder and report results in Appendix C. The encoded 3D scene is represented using three 64x64 feature planes (one for each axis pair XY, YZ, XZ). While many implicit networks have been proposed~\cite{Boulch2022, Chen2022}, the ConvONet has some nice properties. Firstly, most other implicit reconstruction networks assume access to a full pointcloud of the scene, obtained by either sampling points on a mesh or using full 3D scans of scenes. The ConvONet, meanwhile, can reconstruct unseen parts of the scene by using a U-Net architecture over explicit feature grids/planes in the whole scene. Another advantage is that it utilizes feature grids/planes~\cite{Wang2022} is the faster inference time of a ConvONet, which is essential in our case since we need to perform surface rendering at both scene-level and grasp-level in real-robot scenarios.

\paragraph{SE(3) Equivariance}
Many recently proposed methods leverage SO(3) or SE(3) equivariance to make their networks more robust to rotations and translations~\cite{Deng2021,Huang2023}. Since rotational equivariance could be a viable method for obtaining viewpoint-equivariant feature extraction, we also run experiments with an SO(3) equivariant encoder leveraging VectorNeurons~\cite{Deng2021}. Note that SO(3) equivariance cannot work on voxel or feature grids since they do not belong to the rotation group. However, we still want the local feature and scene completion capabilities of ConvONets. To solve this problem, we propose a hybrid VecNeuron-FeatureGrid encoder. The encoder uses a pointcloud input and encodes point-wise 3D equivariant features using a VectorNeuron architecture. These point-wise features are then pooled to obtain SO(3) \textit{invariant} features that can then be projected into the feature planes of ConvONets. Experimental results are in Appendix C.

\subsubsection{Decoder}
\label{method:dec}
The grasp decoder classifies each grasp as successful or unsuccessful, hence predicting grasp quality. For each grasp query, the surface rendering procedure gives us 3D points relevant to the grasp, while the encoded 3D representation provides features corresponding to these points. We transform the 3D surface points (\cref{fig:surf}(b)) to the grasp (gripper) frame and concatenate them to the features. We also concatenate the grasp (gripper) position and orientation to these features. To finally predict the grasp quality, we use a feature point network to process this 3D feature pointcloud. We use two popular network architectures: PointNet~\cite{Qi2017} and DGCNN~\cite{Wang2019}, the comparative results of which are provided in Appendix C. For scene reconstruction, we use the same ResNet-based decoder architecture as ConvONets.

\section{Additional details about experimental setup}\label{appendix:b}


\begin{table}[t!]
\footnotesize
\centering
\captionsetup{justification=centering}
  \caption{List of key network parameters used by \textit{NeuGraspNet}}  \label{params-table}
\begin{adjustbox}{width=\columnwidth,center}
  \begin{tabular}{lc|lc}
    \hline
    parameter & value & parameter & value \\
    \hline
    voxel resolution & 64 & latent dims & 128 \\
    plane resolution & 64 & hidden dims & 256 \\
    U-Net depth & 5 & concat plane feats & True \\
    quality decoder points & 1024 & occupancy loss weight & 2.0 \\
    quality decoder network & PointNet~\cite{Qi2017} & quality loss weight & 1.0 \\
    occupancy decoder ResNets & 5 & local occ loss weight & 1.0\\
    \hline
  \end{tabular}
  \end{adjustbox}
\vspace{-0.6cm}
\end{table}

\subsection{Dataset generation and training}
Unlike semi-implicit methods like \cite{Breyer2020,Jiang2021}, our network is fully implicit in 6 grasp dimensions, which requires a larger number of grasps per scene to learn to discriminate them in SE(3). We use a similar procedure to simulate grasps as in~\cite{Breyer2020, Jiang2021}. We use the scene pointcloud sampled from ground-truth meshes from 33000 simulated scenes from the VGN setup. Pintclouds are cropped to within the 30 cm cube considered for grasping and fed to a GPG sampler~\cite{Gualtieri2016}. We sample a maximum of 60 grasps per scene. For occupancy prediction, we sample 100,000 points and their occupancies using uniform sampling in the scene volume.

As detailed in~\cref{subsec:local}, we also use local supervision for both grasp quality and scene reconstruction. We use the ground truth renderer in simulation to render the local surface points relevant for every grasp in our dataset. As shown in \cref{fig:surf}(a), we use three cameras per grasp (one for every link of the gripper) placed 2.5 cm away from the link and render 64x64 depth images with a 120-degree field of view (FOV). We merge the surface pointclouds from the three cameras to generate our local grasp pointcloud. We filter out points outside the gripper by more than 1cm. We use a maximum of 1024 points per grasp and discard grasps with less than 50 points in their local surface cloud. We also add noise (gamma noise as in \cite{Jiang2021}) to these surface points and store the occupancy values of the points~(\cref{fig:surf}(c)). Training the grasp quality using these noisy pointclouds helps regularize against imperfect neural rendering at test time. Training scene reconstruction with occupancies of these noisy pointclouds helps precisely reconstruct regions close to surfaces. We provide our data generation pipeline as part of our codebase.

We train all our networks on an NVIDIA V100 GPU. Each training run takes about 25 hours to converge. We use 5\% of the data for validation. Learning rates between 1e-4 and 5e-5 work best for all the networks with a batch size of 32 data samples. A list of network parameters used is provided in~\Cref{params-table}.



\begin{table*}[h]
\footnotesize
    \vspace{-0.4cm}
\centering
\captionsetup{justification=centering}
  \caption{Comparative results when using different encoder and decoder architectures}  \label{supp-ablat-table}
  \begin{tabular}{@{}lcc|cc|cc@{}}
    \hline
     {} & \multicolumn{6}{c}{Pile scenes}\\
    \hline
    {} & \multicolumn{2}{c}{Fixed Top View}&\multicolumn{2}{c}{Random View}&\multicolumn{2}{c}{Hard View}\\
    \cline{2-7}
    Method & GSR (\%) & DR (\%) & GSR (\%)& DR (\%)& GSR (\%)& DR (\%) \\
    \hline
    NeuGraspNet (3D Enc+PointNet Dec) & \textbf{86.51 $\pm$ 1.42} & \textbf{83.52 $\pm$ 2.24} & \textbf{85.05 $\pm$ 1.25} & \textbf{84.37 $\pm$ 1.52} & \textbf{73.95 $\pm$ 1.26} & \textbf{70.67 $\pm$ 1.69} \\
\hline
\hline

    PointNet Encoder & 79.26 $\pm$ 3.36 & 78.98 $\pm$ 2.61 & 77.82 $\pm$ 1.60 & 78.45 $\pm$ 2.72 & 66.64 $\pm$ 1.85 & 64.50 $\pm$ 1.82 \\

    VectorNeuron Encoder & 79.21 $\pm$ 1.87 & 77.82 $\pm$ 2.92 & 77.28 $\pm$ 1.32 & 77.60 $\pm$ 1.88 & 63.57 $\pm$ 2.40 & 60.05 $\pm$ 3.30 \\

    DGCNN Decoder & 84.91 $\pm$ 1.03 & 83.14 $\pm$ 1.54 & 81.80 $\pm$ 1.98 & 80.68 $\pm$ 2.14 & 68.63 $\pm$ 1.46 & 65.29 $\pm$ 2.45 \\
        \hline
     {} & \multicolumn{6}{c}{Packed scenes}\\
    \hline
    {} & \multicolumn{2}{c}{Fixed Top View}&\multicolumn{2}{c}{Random View}&\multicolumn{2}{c}{Hard View}\\
    \cline{2-7}
    Method & GSR (\%) & DR (\%) & GSR (\%)& DR (\%)& GSR (\%)& DR (\%) \\
    \hline
        NeuGraspNet (3D Enc+PointNet Dec) &  \textbf{97.65 $\pm$ 0.92} & \textbf{93.16 $\pm$ 1.48} & \textbf{92.49 $\pm$ 1.41} & \textbf{91.74 $\pm$ 1.24} & \textbf{78.76 $\pm$ 1.89} & \textbf{82.80 $\pm$ 1.50} \\
\hline
\hline
    PointNet Encoder & 95.15 $\pm$ 1.86 & 90.40 $\pm$ 0.88 & 89.80 $\pm$ 1.94 & 88.97 $\pm$ 2.04 & 73.86 $\pm$ 1.64 & 77.15 $\pm$ 2.16 \\
     
    VectorNeuron Encoder & 94.75 $\pm$ 0.63 & 90.61 $\pm$ 1.45 & 89.16 $\pm$ 2.77 & 89.10 $\pm$ 1.86 & 77.29 $\pm$ 2.84 & 80.05 $\pm$ 2.87 \\

    DGCNN Decoder & 95.61 $\pm$ 0.73 & 93.01 $\pm$ 1.95 & 90.62 $\pm$ 1.83 & 90.78 $\pm$ 1.32 & 74.69 $\pm$ 2.58 & 79.94 $\pm$ 2.41 \\

    \hline
  \end{tabular}
\end{table*}

\begin{figure*}[h]
     \centering
     \begin{subfigure}[b]{0.475\textwidth}
         \centering
         \includegraphics[width=\textwidth,trim={0cm 5cm 0cm 3.2cm},clip]{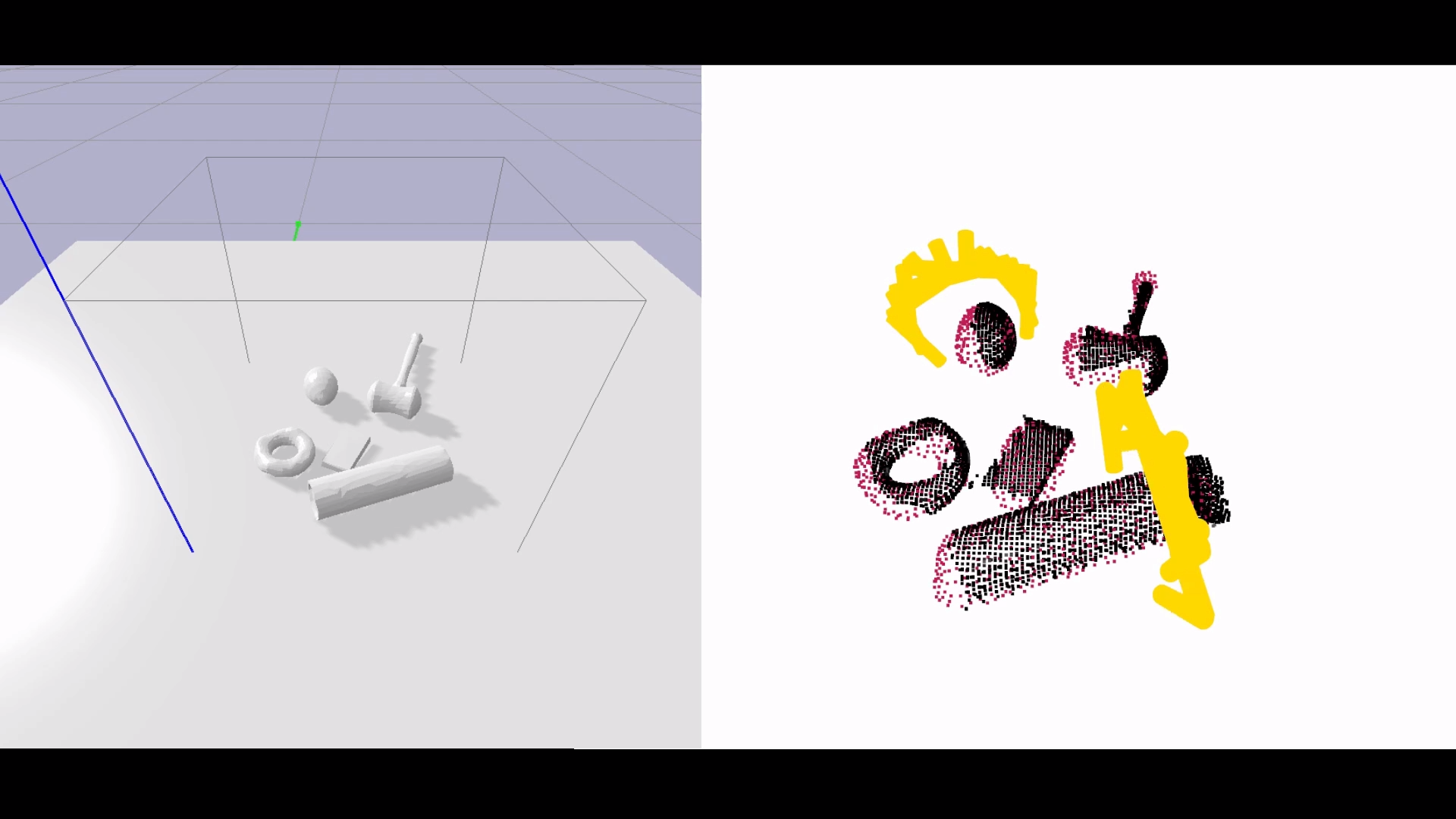}
        \vspace{-0.4cm}
         \caption{}
     \end{subfigure}
     \begin{subfigure}[b]{0.475\textwidth}
         \centering
         \includegraphics[width=\textwidth,trim={0cm 5cm 0cm 3.2cm},clip]{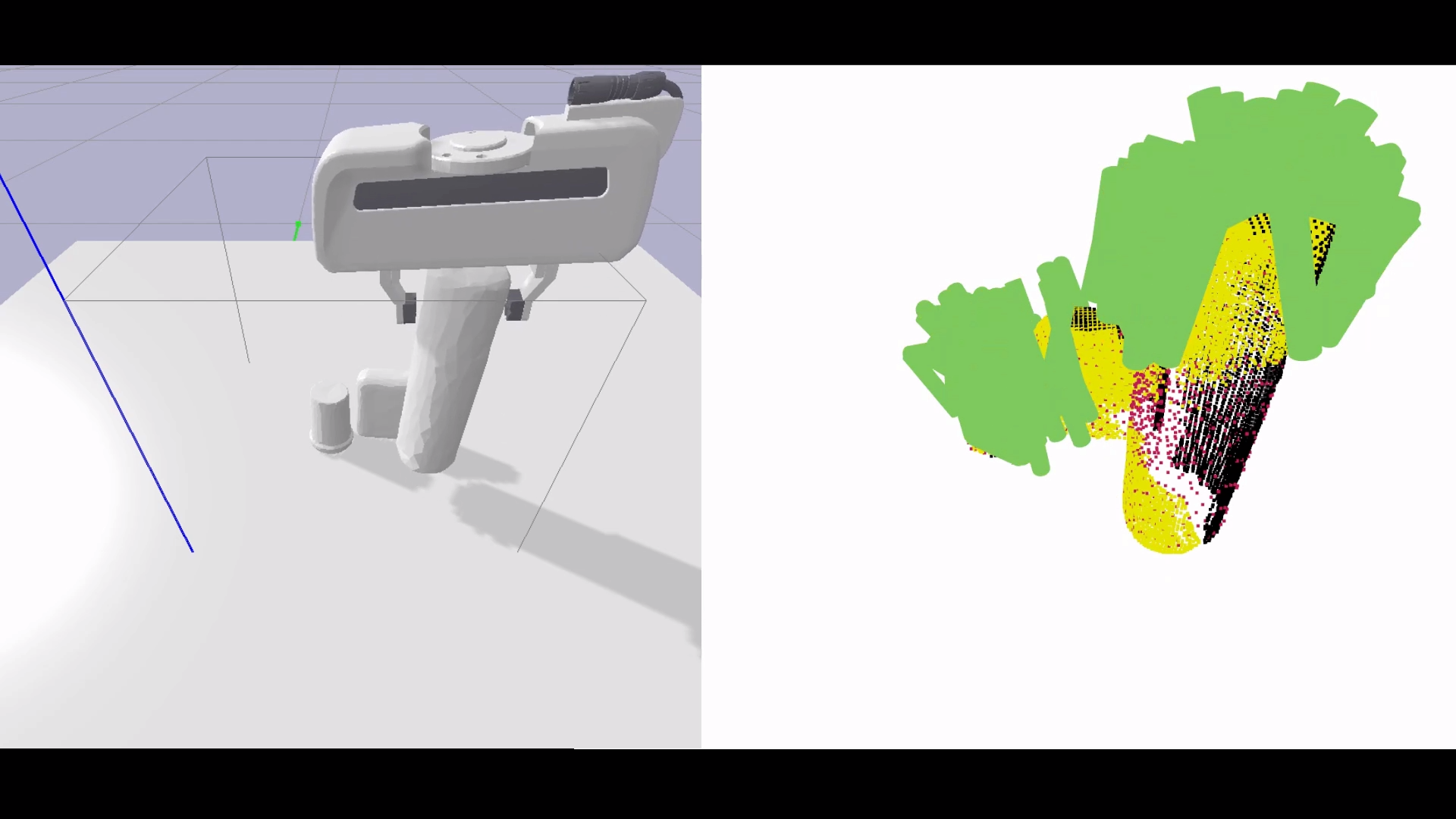}
        \vspace{-0.4cm}
         \caption{}
     \end{subfigure}
    \caption{Experiments on (a) pile and (b) packed scenes from the VGN \cite{Breyer2020} simulation benchmark. We visualize the input scene pointcloud (black), our reconstructed pointcloud (red), candidate grasps (yellow), and detected high-quality grasps (green).}
    \label{fig:sim_exp}
\end{figure*}

\subsection{Neural rendering}
For the neural rendering of the implicit geometry at inference time, we use a two-stage depth prediction method as used in \cite{niemeyer2020differentiable, Oechsle2021}. The first stage proposes points along each ray originating from the camera and checks their occupancy using the occupancy network. The first point along a ray for which the occupancy changes from free space to occupied space is a rough estimation of the surface point. The second stage uses a secant search root-finding method along the smaller ray segment around the roughly estimated surface point. This search gives us the precise surface point.

For scene-level neural rendering, we use six cameras placed at a 60-degree elevation around the scene~(\cref{fig:grasp_detec_steps}(b)). Each camera renders 64x64 images with a 90-degree FOV. We propose 256 proposal points along each ray at a maximum distance of 3$\times$(size of the scene) = 90 cm, giving us a resolution of 0.0035 m. We then perform 8 secant search steps to refine the surface point estimation further.

For local grasp-level neural rendering, we use the same camera placement and settings used for the local supervision ground-truth rendering (64x64 images and 120-degree FOV). We propose 15 proposal points along each ray at a maximum distance of 2$\times$(gripper width) = 9 cm. This gives us a resolution of 0.006 m. We then perform 8 secant search steps to refine the surface point estimation further to arrive at the rendered surface cloud visualized in \cref{fig:surf}(b).

\section{Additional experimental results}\label{appendix:c}

\begin{figure*}[ht!]
\centering
  \includegraphics[width=0.999\linewidth]{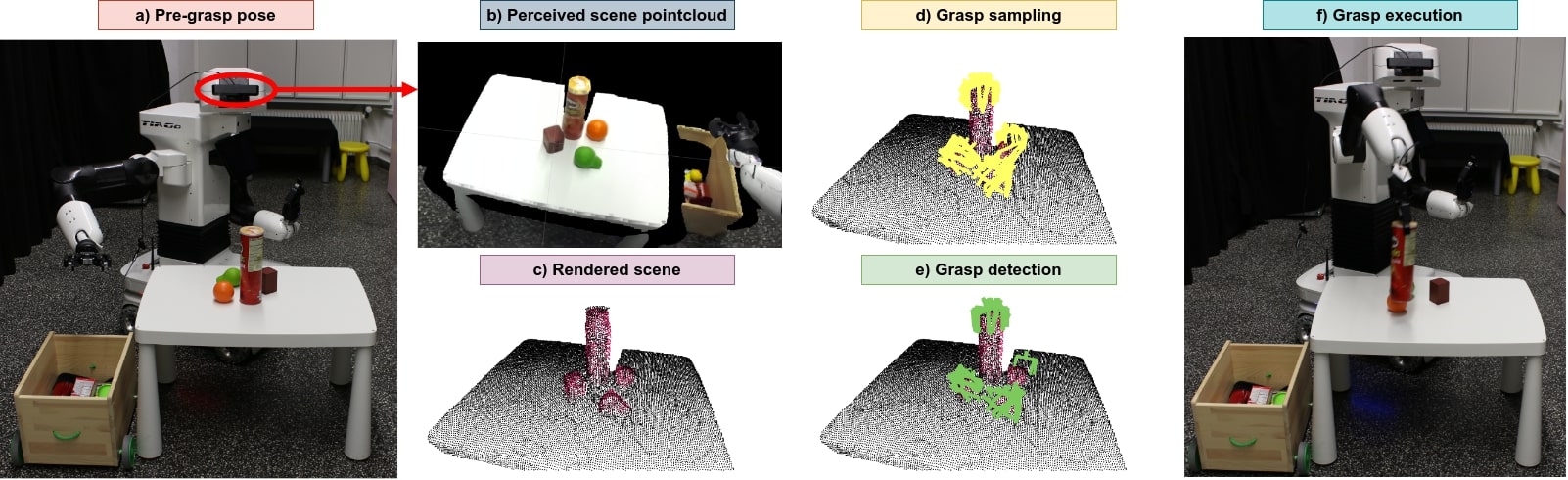}
  \caption{Real-world evaluation of NeuGraspNet with a mobile manipulator robot. The robot starts from a pre-grasp pose (a), and perceives the scene pointcloud (b). NeuGraspNet renders and completes the scene (c), runs 6DoF grasp sampling (d), and detects high-quality grasps (e). The robot executes grasps with classical motion planning, collecting objects in the box (f).}
  \label{fig:realrobot}
\end{figure*}

\begin{figure*}[h]
     \centering
     \begin{subfigure}[b]{0.175\textwidth}
         \centering
         \includegraphics[width=\textwidth,trim={5cm 5cm 5cm 8cm},clip]{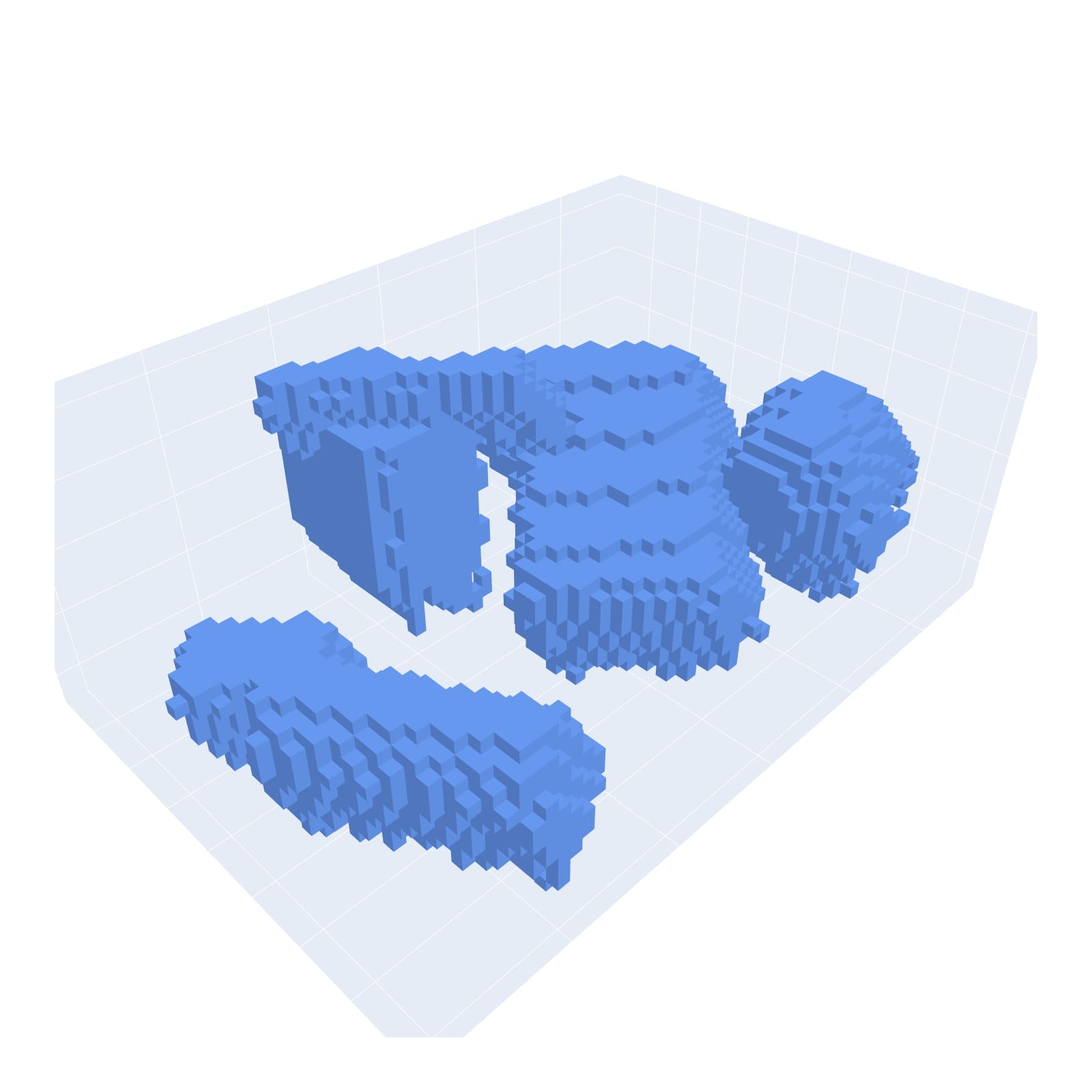}
        \vspace{-0.4cm}
     \end{subfigure}
     \begin{subfigure}[b]{0.175\textwidth}
         \centering
         \includegraphics[width=\textwidth,trim={5cm 5cm 5cm 8cm},clip]{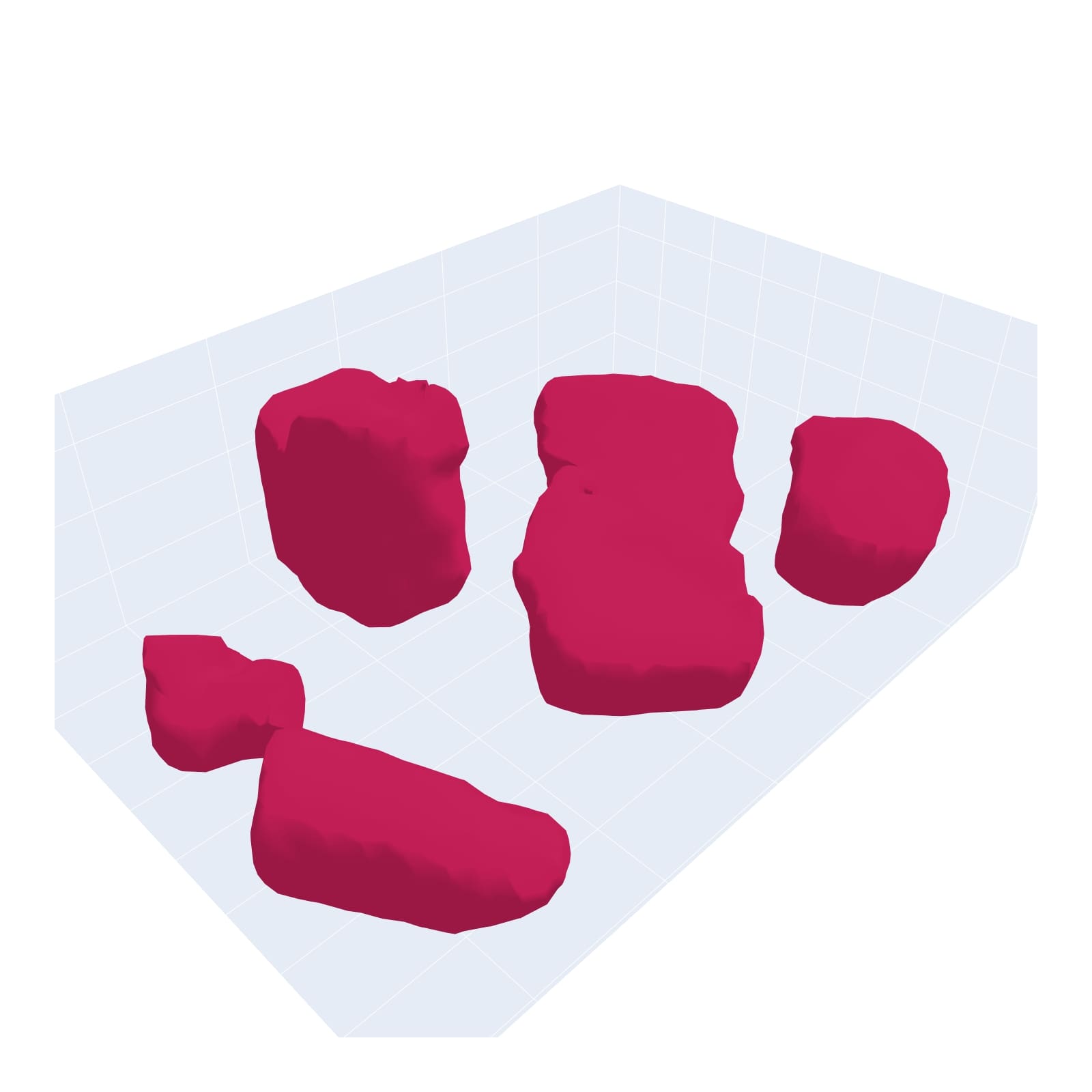}
        \vspace{-0.4cm}
     \end{subfigure}
      \begin{subfigure}[b]{0.175\textwidth}
         \centering
         \includegraphics[width=\textwidth,trim={5cm 5cm 5cm 8cm},clip]{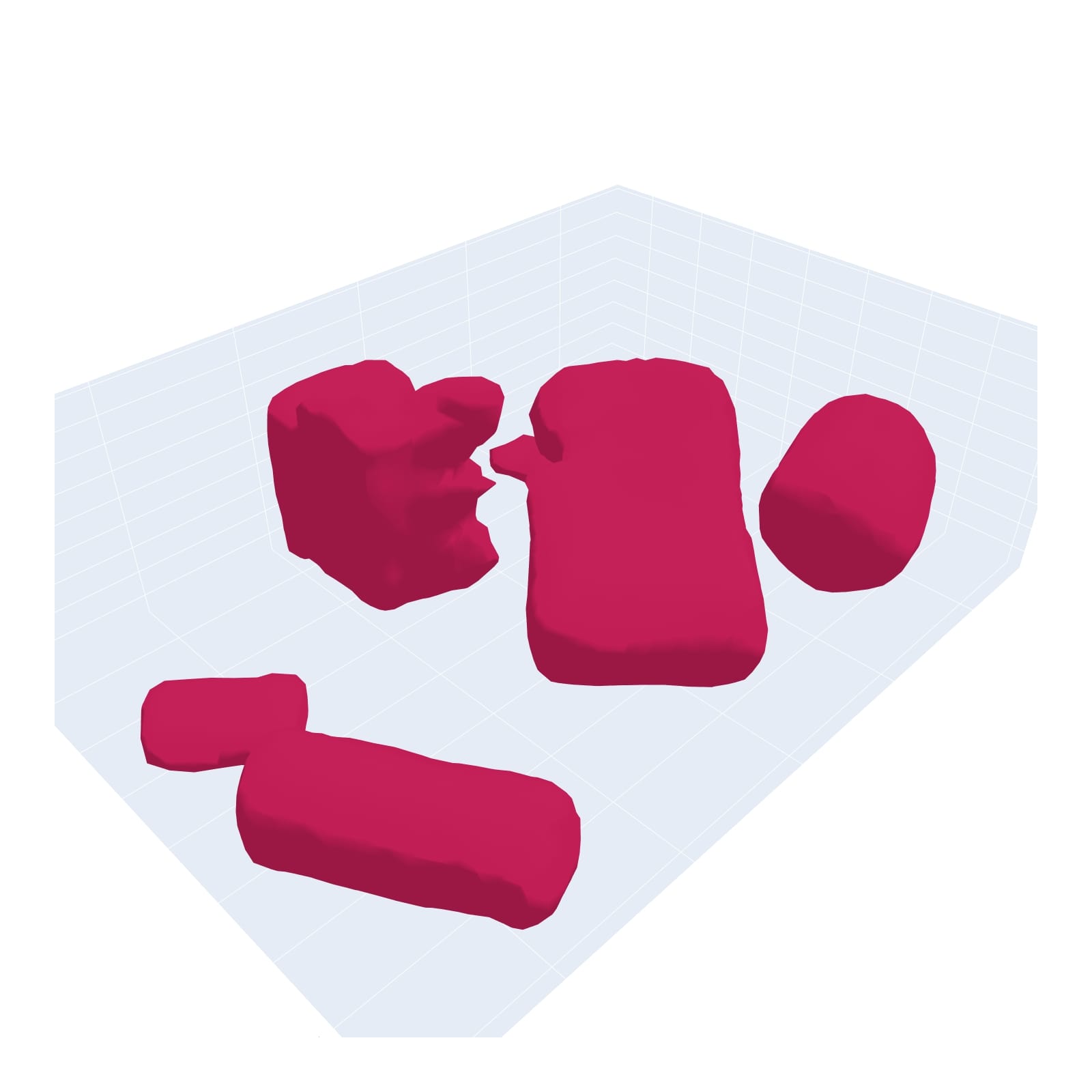}
        \vspace{-0.4cm}
     \end{subfigure}
      \vspace{-0.1cm}
      
      \begin{subfigure}[b]{0.175\textwidth}
         \centering
         \includegraphics[width=\textwidth,trim={2cm 2cm 5cm 4cm},clip]{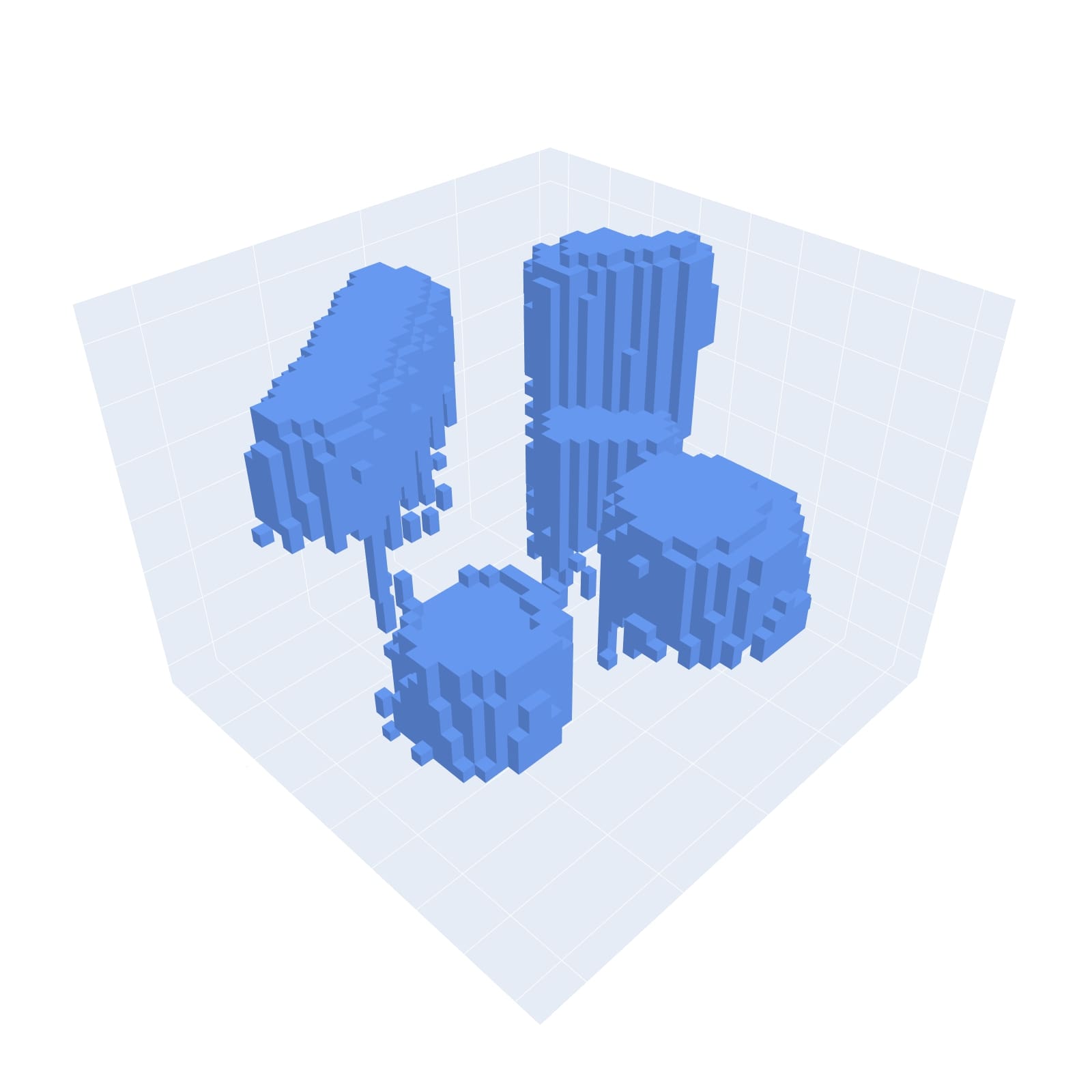}
        \vspace{-0.4cm}
         \caption{Input TSDF}
     \end{subfigure}
     \begin{subfigure}[b]{0.175\textwidth}
         \centering
         \includegraphics[width=\textwidth,trim={5cm 5cm 5cm 7cm},clip]{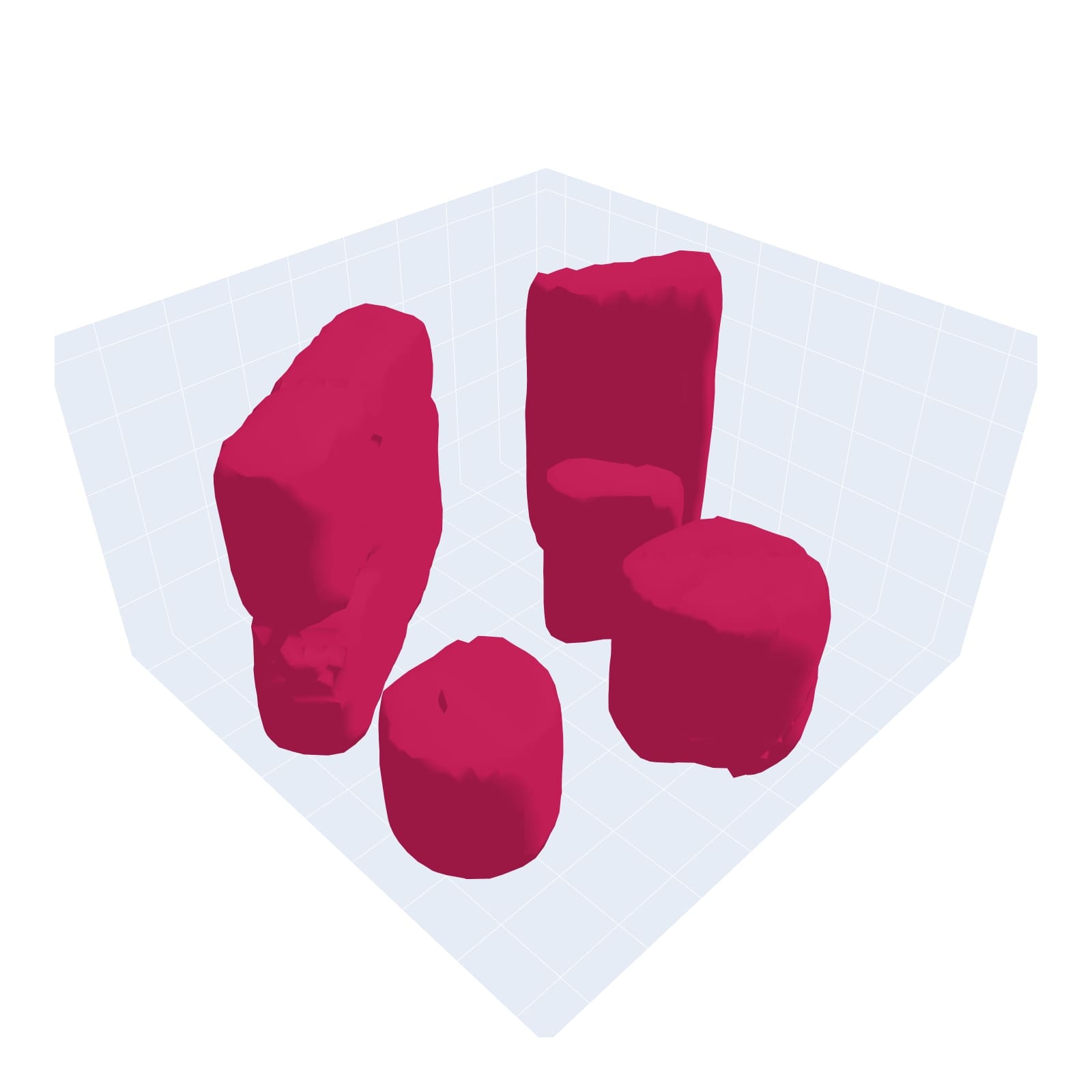}
        \vspace{-0.4cm}
         \caption{No-loc-occupancy}
     \end{subfigure}
      \begin{subfigure}[b]{0.175\textwidth}
         \centering
         \includegraphics[width=\textwidth,trim={5cm 5cm 5cm 7cm},clip]{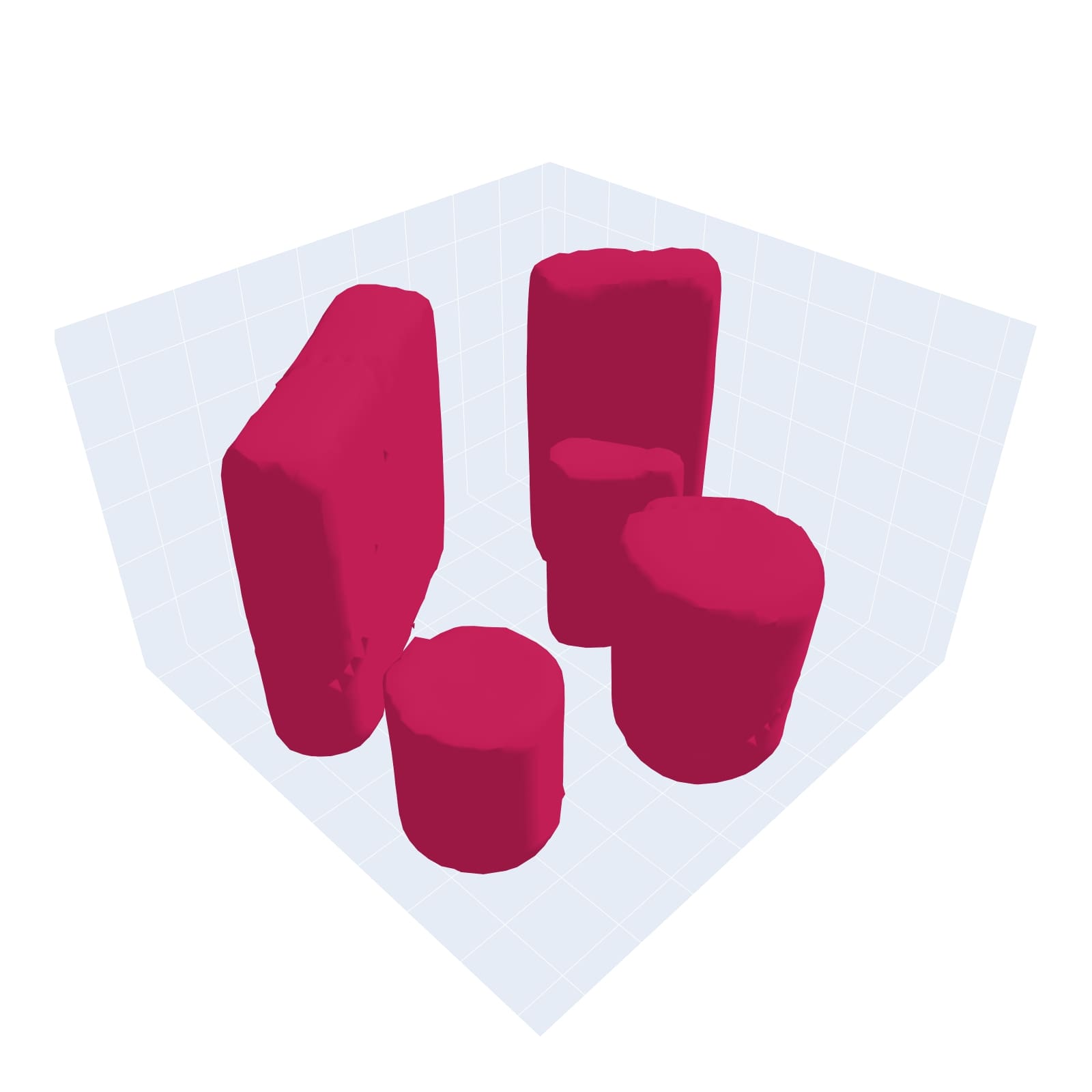}
        \vspace{-0.4cm}
         \caption{NeuGraspNet}
     \end{subfigure}
    \caption{Example scene reconstructions in a `pile' (top) and a `packed' (bottom) scene. Networks trained \textit{with} local occupancy supervision (c) are able to finely reconstruct the local surface.}
    \label{fig:scene-recon}
\end{figure*}

\subsection{Additional comparisons/ablations}

In \cref{supp-ablat-table}, we compare the results of NeuGraspNet against other encoder and decoder architectures as mentioned in \cref{appendix:a}. Notably, the network performance drops when an input pointcloud is used instead of the TSDF grid and the 3D Encoder. Thus, both the PointNet and our adapted VectorNeuron encoder versions of the network perform worse. In pile scenes, the VectorNeuron encoder does not provide any notable benefits but for packed scenes, we see that the VectorNeuron version performs better than the PointNet encoder with hard viewpoints. In the case of the grasp quality decoder, the DGCNN point decoder performs slightly worse than the PointNet decoder used in the final NeuGraspNet model. A possible explanation is that the DGCNN network could struggle to fit the data when the previously encoded latent features of the pointset change during training as the occupancy predictions change. These feature changes could cause issues in the dynamic knn feature graph generation performed by DGCNN.

\subsection{Scene reconstruction}

We report the scene reconstruction results of NeuGraspNet qualitatively in~\cref{fig:scene-recon} and quantitatively in~\cref{scene-recon-table}. Though scene reconstruction is not the primary goal of our work, we run experiments to note how accurate the scene reconstruction needs to be to achieve the grasping results of NeuGraspNet. Moreover, we check the effect of our local occupancy supervision in `pile' and `packed' scenes. We reconstruct the scene mesh by querying the network over the volume using the Multiresolution IsoSurface Extraction (MISE) method from \cite{Mescheder2019}. We report the Intersection over Union (IoU) between the ground truth mesh and the reconstructed mesh in~\cref{scene-recon-table}. Notably, even though we use TSDF grids from randomized viewpoints in our problem setup (making reconstruction much more challenging than in~\cite{Jiang2021}), the network still performs well. Consistent with the grasping results, local occupancy supervision is beneficial in `packed' scenes leading to a higher IoU, but has a smaller effect in `pile' scenes~(\cref{scene-recon-table}).

\begin{table}[t]
\centering
\captionsetup{justification=centering}
  \caption{Scene reconstruction results}  \label{scene-recon-table}
  \begin{tabular}{@{}lc|lc@{}}
    \hline
     \multicolumn{2}{c}{Pile scenes} & \multicolumn{2}{c}{Packed scenes}\\
    \hline
    Method & IoU (\%) &  Method & IoU (\%) \\
    \hline
    NeuGraspNet & 73.22 & NeuGraspNet & 89.39 \\
    No-local-occ & 73.48 & No-local-occ & 87.16 \\
    \hline
  \end{tabular}
\end{table}



\begin{figure*}[ht!]
    \centering
    \includegraphics[width=0.8\textwidth]{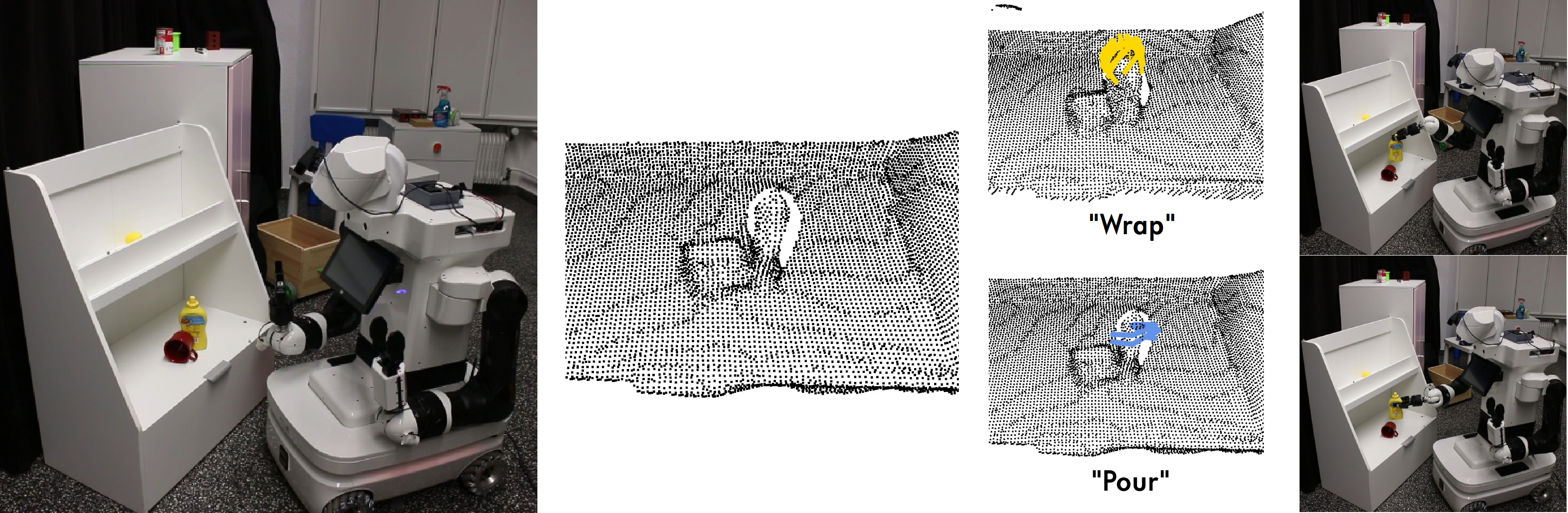}
    \caption{Example grasp affordance prediction in a mobile manipulator grasping scenario.}
    \label{fig:real_aff}
\end{figure*}

\subsection{Real-world experiments}

To demonstrate the applicability of our method in the real-world, we perform grasping experiments with a TIAGo++ mobile manipulator with an omnidirectional base. We create several mobile manipulation scenarios such as grasping from a shelf, from tables, from the top of cupboards etc. The viewing angle of the robot is randomized by moving the base of the robot and moving the torso joint to a random position. The head of the robot is pointed towards the target area where objects need to be grasped, as visualized in~\Cref{fig:realrobot}. For depth estimation, we use a ZED2 stereo camera mounted over the head of the robot. We use the same grasp sampling (GPG) and rendering settings as in the simulation experiments. Unlike the simulations, we use a Robotiq 2F-85 adaptive gripper for grasping objects. Thus, we generate another dataset from simulation using our dataset generation strategy with a simulated Robotiq gripper. To ensure robustness for sim-to-real transfer, we add more noise to the simulated depth cameras at both the scene and grasp rendering levels. This noise also helps generalize the network to cases where the support surface it lies on is not perfectly even eg. natural ground.  We use the MoveIt planning library to plan the robot arm's movements. As grasp objects, we use objects from the YCB dataset~\cite{calli2015benchmarking}. For affordance prediction, we further add objects from the categories considered in our grasp affordance dataset, namely `Knife’, `Scissors’, `Mug’, `Bottle’, `Bag’, `Earphone’, and `Hat’. FOr demonstrating the ability of NeuGraspNet in task-oriented grasping, we show grasp detection based on user-queried affordances. Example grasp affordance predictions are visualized in~\Cref{fig:real_aff}. 
Video demonstrations of the real-world grasp execution are provided at \url{https://sites.google.com/view/neugraspnet}.


\end{document}